\documentclass[5p,twocolumn]{elsarticle}

\makeatletter
\def\ps@pprintTitle{
  \let\@oddhead\@empty
  \let\@evenhead\@empty
  \def\@oddfoot{\reset@font\hfil\thepage\hfil}
  \let\@evenfoot\@oddfoot
}
\makeatother

\usepackage{hyperref}
\usepackage[labelfont=bf,font=normalsize]{caption}
\usepackage{array} 
\usepackage{bm} 
\usepackage{arydshln} 
\usepackage{makecell} 
\usepackage{amsmath} 
\usepackage{subcaption} 
\usepackage{multirow}

\begin{document}
\begin{frontmatter}
    \title{\textit{ISLA}: Ensemble learning and domain adaptation for generalizable acute ischemic stroke lesion segmentation in diffusion MRI with clinical validation}
    \author[1]{Vincent Roca\corref{cor1}}
    \author[2]{Hakim Daouairi}
    \author[2]{Sevak Yepremian}
    \author[3]{Hilde Henon}
    \author[4]{Nicolas Martinage}
    \author[3,5]{Laurent Puy}
    \author[1,2,5]{Grégory Kuchcinski}
    \author[1,2,5]{Martin Bretzner}
    \author[1,4,5]{Renaud Lopes}
    \cortext[cor1]{Corresponding author at: Lille, 59037 France. E-mail address: vincent.roca@chu-lille.fr}
    \address[1]{Univ. Lille, CNRS, Inserm, CHU Lille, Institut Pasteur de Lille, US 41 - UAR 2014 - PLBS, F-59000 Lille, France}
    \address[2]{CHU Lille, Département de Neuroradiologie, F-59000 Lille, France}
    \address[3]{CHU Lille, Département de Neurologie vasculaire, F-59000 Lille, France}
    \address[4]{CHU Lille, Département de Médecine Nucléaire, F-59000 Lille, France}
    \address[5]{Univ. Lille, Inserm, CHU Lille, U1172 - Lille Neurosciences \& Cognition, F-59000 Lille, France}
    
    \begin{abstract}
        Accurate segmentation of acute ischemic stroke (AIS) lesions on diffusion-weighted MRI is essential for clinical decision-making. Deep learning methods offer a promising approach to automating this task. However, they often generalize poorly to unseen clinical sites, and few AIS lesion segmentation methods have been externally validated. In this study, we propose \textit{ISLA} (Ischemic Stroke Lesion Analyzer), a framework for AIS lesion segmentation built on the design principles of \textit{nnU-Net}. \textit{ISLA} combines attention modules, deep supervision, and ensemble learning across different input resolutions, convolutional block types, and loss functions. We also introduce \textit{ISLA-MT}, a variant that leverages unlabeled target-domain images through semi-supervised learning. We trained both approaches on the SOOP and ISLES 2022 datasets and evaluated on a held-out internal validation set of 253 cases, as well as on 590 cases from an independent multicenter clinical cohort. We compared them with three state-of-the-art pre-trained models and with \textit{nnU-Net} trained on the same data. \textit{ISLA} outperformed the pre-trained models on the internal validation set, while both \textit{ISLA} and \textit{ISLA-MT} outperformed them on the external cohort. Although \textit{nnU-Net} performed slightly better than \textit{ISLA} internally, both \textit{ISLA} variants demonstrated better generalization to the external cohort. Furthermore, \textit{ISLA-MT} significantly improved performance over \textit{ISLA} on the external cohort. The proposed models performed particularly well on low-contrast lesions, and qualitative error analyses suggested greater robustness to acquisition artifacts. Both models are fast ($\sim$ 5 s per case), and their code and weights are publicly available at \url{https://github.com/RocaVincent/isla_segmentation}. 
    \end{abstract}

    
\end{frontmatter}

\section{Introduction}
Stroke is a major cause of mortality and long-term disability, with ischemic strokes accounting for approximately 87\% of all cases \citep{Capirossi2023}. Accurate delineation of ischemic lesions in acute MRI is critical for clinical decision-making \citep{Cimflova2021}. Although manual delineation by experts remains the gold standard, it is time-consuming and labor-intensive.

Automatic segmentation of acute ischemic stroke (AIS) lesions has therefore become an active area of research. Supported by the availability of expert-annotated, multicenter MRI datasets such as \textit{ISLES} 2022 \citep{HernandezPetzsche2022}, deep learning has emerged as a powerful tool for MRI-based AIS lesion segmentation \citep{Abbasi2023,Luo2024}, most commonly using diffusion-weighted imaging \citep{Kamel2025}.

Among the proposed methods, U-Net variants dominate the field, with design variations including loss functions, deep supervision, residual connections, and attention mechanisms \citep{Abbasi2023,Baaklini2025,Luo2024}. However, comparing approaches remains challenging because most implementations are not publicly accessible \citep{Baaklini2025}. Consequently, the optimal configuration for AIS lesion segmentation remains unclear.

Besides, \textit{nnU-Net}---a self-configuring U-Net framework \citep{Isensee2021}---has consistently outperformed standard U-Net variants and has been widely adopted across diverse medical image segmentation tasks \citep{Antonelli2022,Campello2021,Wasserthal2023}. One of its main strengths is its ability to adapt automatically to the characteristics of the input dataset, particularly when handling 3D anisotropic images, which are common in AIS MRI. Despite its strong performance and widespread use, \textit{nnU-Net} has been only sparsely investigated for AIS lesion segmentation.

Ensemble learning, a key component of \textit{nnU-Net}, is commonly used to improve robustness and reduce overfitting. By aggregating predictions from multiple independently trained models, ensemble methods have demonstrated consistent performance gains across a range of medical image segmentation tasks \citep{Li2022,Wiltgen2024}, including AIS lesion segmentation \citep{Jeong2024,delaRosa2025}.

Beyond performance on internal test sets, evaluation on external datasets is essential for assessing real-world applicability. Variability in scanner hardware, acquisition protocols, and site-specific practices introduces substantial distribution shifts across centers, making model generalizability a persistent challenge \citep{Luo2024,ZafariGhadim2024,Vorberg2025}. Recent meta-analyses have highlighted the limited use of external validation in stroke detection and segmentation studies \citep{Bojsen2024,Baaklini2025}. In this context, domain adaptation has emerged as a promising strategy to enhance the cross-site robustness of deep learning models for AIS lesion segmentation \citep{Abbasi2023,Luo2024}.

In this study, we propose \textit{ISLA} (Ischemic Stroke Lesion Analyzer), a novel framework for AIS lesion segmentation from diffusion-weighted imaging. Building on design principles from \textit{nnU-Net}, \textit{ISLA} integrates attention mechanisms and multiple model variations combined through ensemble learning, including different input resolutions, convolutional block types (standard or residual), and loss functions. Furthermore, we introduce \textit{ISLA-MT}, a variant incorporating the Mean Teacher (MT) framework for unsupervised domain adaptation (UDA) to improve generalization to external datasets.

The proposed models were trained on public datasets and evaluated on an internal validation dataset and an independent multicenter clinical cohort of 590 patients. Comparative experiments against three state-of-the-art AIS lesion segmentation methods, as well as \textit{nnU-Net}, demonstrate the superior performance of \textit{ISLA}, with further generalization improvements achieved using \textit{ISLA-MT}. The code and trained model weights are available at \url{https://github.com/RocaVincent/isla_segmentation}.

\section{Related works}
This section reviews technical components that have been adopted in recent years in medical image segmentation---with a particular emphasis on AIS lesion segmentation---and that we incorporated into the \textit{ISLA} framework.

\subsection{Loss functions}
\label{relWorks:loss_fn}
A common challenge in medical image segmentation is the severe class imbalance between foreground and background voxels. To account for this, Dice loss \citep{Milletari2016} has been widely adopted for segmentation of stroke lesions \citep{Wong2022,Yu2023,Li2024}. To provide stronger and more stable gradients—particularly during the early stages of training—it is often combined with cross-entropy loss \citep{Jeong2024,Cao2025,Chalcroft2025}.

To further address class imbalance, \citet{Yeung2022} introduced Unified Focal Loss (UFL), a generalization of cross-entropy-based and Dice loss functions. In comparison against six widely used losses across five medical image segmentations tasks---including MRI-based brain tumor segmentation---\citeauthor{Yeung2022} consistently reported higher Dice coefficients with UFL.

\subsection{U-Net variations}
\textbf{Deep supervision.} Deep supervision introduces additional loss terms at intermediate outputs of deep neural networks, encouraging hidden layers to learn more task-discriminative representations \citep{Lee2015}. It has been successfully applied to MRI-based stroke segmentation \citep{Liu2021,Sheng2022,Maqsood2025}. Notably, \citet{Gomez2023} reported improved performance for AIS lesion segmentation when using deep supervision.

\textbf{Residual connections.} By adding the input directly to the output within convolutional blocks \citep{He2016}, residual connections allow gradients to bypass convolutions, helping to mitigate vanishing gradients. These connections are widely used in U-Net-based architectures for AIS lesion segmentation \citep{Liu2020,Clerigues2020,Alis2021,Gheibi2023}.

\textbf{Attention mechanisms.} Attention mechanisms encourage convolutional neural networks to focus on the most informative image features. They are commonly categorized into spatial attention, which highlights specific image regions \citep{Schlemper2019}, and channel attention, which emphasizes informative feature channels \citep{Hu2020}. They have been widely explored for stroke lesion segmentation, including spatial attention \citep{Liu2020,Yu2020,Yu2021,NazariFarsani2023}, channel attention \citep{Abramova2021}, and hybrid mechanisms combining both mechanisms \citep{Liu2021,Alshehri2023,Rahman2025}.

\subsection{\textit{nnU-Net}}
\textit{nnU-Net} is a framework for deep learning-based medical image segmentation \citep{Isensee2021}. Based on the characteristics of the training dataset and cross-validation results, it automatically configures preprocessing steps, network architecture, and post-processing operations.

This self-configuring strategy enables adaptation to a wide range of segmentation tasks. Combined with the strong performance demonstrated across numerous medical image segmentation benchmarks \citep{Campello2021,Antonelli2022}, \textit{nnU-Net} has become the de facto standard for supervised medical image segmentation. Notably, it outperformed several alternative deep learning architectures for MRI-based AIS lesion segmentation \citep{Jeong2024,Kamel2025}.

\subsection{Image resampling}
While medical imaging datasets often contain heterogeneous spatial resolutions, convolutional neural networks typically require prior resampling to ensure consistent input spacing \citep{Isensee2021}. In the context of MRI-based AIS lesion segmentation, images frequently exhibit anisotropic resolution, with high in-plane (axial) resolution and relatively large slice thickness \citep{Liu2021,Moon2022,Ryu2025}.

Different strategies have been adopted to deal with resolution heterogeneity and anisotropy. Some models operate in 2D by excluding the weakly resolved axis \citep{Wong2022,Moon2022}, but this approach prevents the exploitation of full 3D contextual information \citep{Li2024}. Other methods rely on resampling to a common isotropic 3D resolution \citep{Liu2021,Li2024,Chalcroft2025,delaRosa2025}, although this can introduce substantial interpolation artifacts for highly anisotropic images.

Alternatively, some 3D models avoid interpolation by keeping the weak resolution unchanged \citep{Wei2022,Bridge2022}. In \textit{nnU-Net}, interpolation is limited by explicitly preserving anisotropy in the target spacing \citep{Isensee2021}.

\subsection{Ensemble learning}
Ensemble methods combine the predictions of multiple models to improve robustness, reduce overfitting, and enhance prediction stability. In MRI-based segmentation, previous studies have aggregated the probability outputs of different U-Net models to obtain more accurate segmentations \citep{Li2022,Wiltgen2024}. Similar performance improvements have also been reported for AIS lesion segmentation \citep{Jeong2024,delaRosa2025}. In these studies, architectural variations among the ensembled models increased prediction diversity, which may further improve generalization performance.

\subsection{Unsupervised domain adaptation}
Systematic differences in MRI data arising from scanner hardware, imaging parameters, and acquisition protocols have long been documented \citep{Abbasi2024}. To mitigate performance degradation on external test sets, domain generalization methods aim to improve model robustness across unseen domains \citep{Dinsdale2021,Yu2023}. However, these approaches typically require training on multiple diverse source domains with balanced label distributions \citep{Dinsdale2021,Roca2025} and are not designed to leverage target-domain data.

In contrast, domain adaptation exploits target-domain data to improve model robustness. Previous studies have shown that supervised domain adaptation can improve generalization in MRI-based AIS lesion segmentation \citep{Alis2021,Ryu2025}, but these methods require labeled data from the target domain. UDA, by contrast, leverages unlabeled target data \citep{Wang2023}. In the context of MRI-based segmentation of spinal cord grey matter, \citet{Perone2019} adapted the MT framework for UDA and demonstrated improved segmentation accuracy on the target-center test set. Although \citet{Cui2019} applied MT to ischemic stroke lesion segmentation, their objective was to increase the effective training set size rather than perform UDA, as all MRI scans were acquired on the same scanner. To our knowledge, no previous study has applied MT-based UDA to stroke lesion segmentation.

\section{Proposed model}
\begin{figure*}
    \centering
    \includegraphics[scale=0.7]{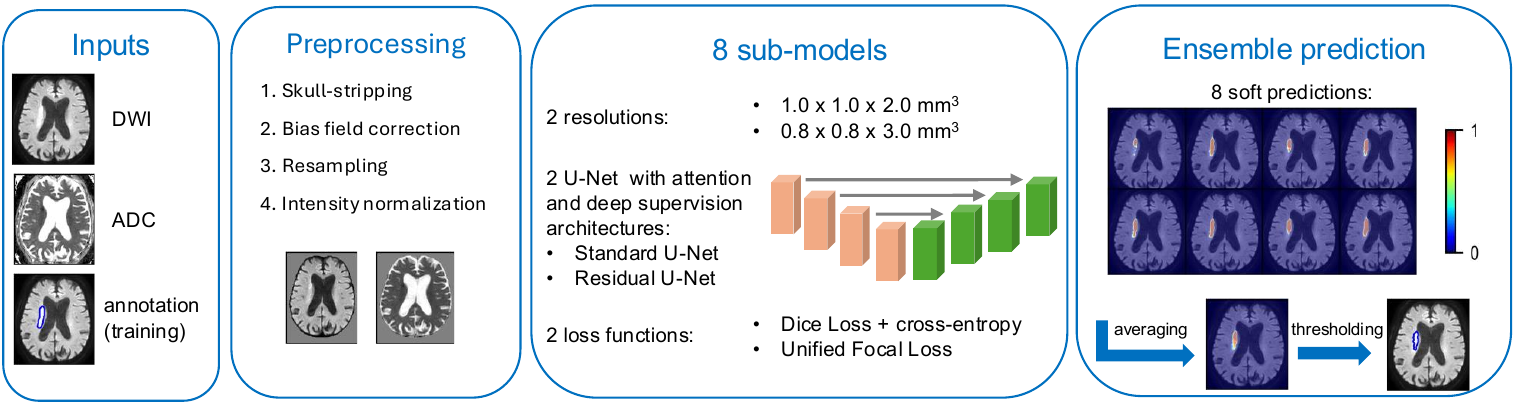}
    \caption{\textbf{Overview of the \textit{ISLA} framework.} 8 sub-models are trained independently on the preprocessed images. The sub-models differ in input resolution, convolutional block design, and loss function. Final \textit{ISLA} predictions are obtained by averaging the predictions of all sub-models.}
    \label{fig:isla_framework}
\end{figure*}

\textit{ISLA} takes a trace diffusion-weighted image (DWI) and the corresponding apparent diffusion coefficient map (ADC) as inputs for AIS lesion segmentation (Figure \ref{fig:isla_framework}). The complete training and inference pipeline are available at \url{https://github.com/RocaVincent/isla_segmentation}.

\subsection{Preprocessing}
\label{meth:preprocessing}
To reduce technical variability and facilitate lesion segmentation, several preprocessing steps are applied before the images are fed to the model.

First, a brain mask is estimated from the DWI using SynthStrip \citep{Hoopes2022} to skull-strip DWI and ADC. Brain DWI intensities are then corrected for bias field inhomogeneity using N4ITK \citep{Tustison2010}.

The MR images are subsequently resampled to a common resolution while accounting for their anisotropic voxel spacing, which is typical of clinical MRI acquisitions. Specifically, in-plane resampling is performed using third-order spline interpolation. For out-of-plane resampling, nearest-neighbor interpolation is used when the corresponding upsampling factor is greater than or equal to 1.5 to prevent interpolation artifacts \citep{Isensee2021}; otherwise, third-order spline interpolation is applied. After resampling, images are padded and/or cropped to predefined target dimensions such that the brain is centered within the resulting image. To account for voxel-spacing heterogeneity in multicenter MRI datasets, \textit{ISLA} incorporates two target resolutions through ensemble learning: 1.0×1.0×2.0 mm$^3$ and 0.8×0.8×3.0 mm$^3$.

Finally, brain intensities are normalized using z-score normalization (separately for DWI and ADC) and clipped to the range $[-5,5]$.

\subsection{Baseline U-Net}
\label{meth:baseline_unet}
\begin{figure*}
    \centering
    \includegraphics[scale=0.52]{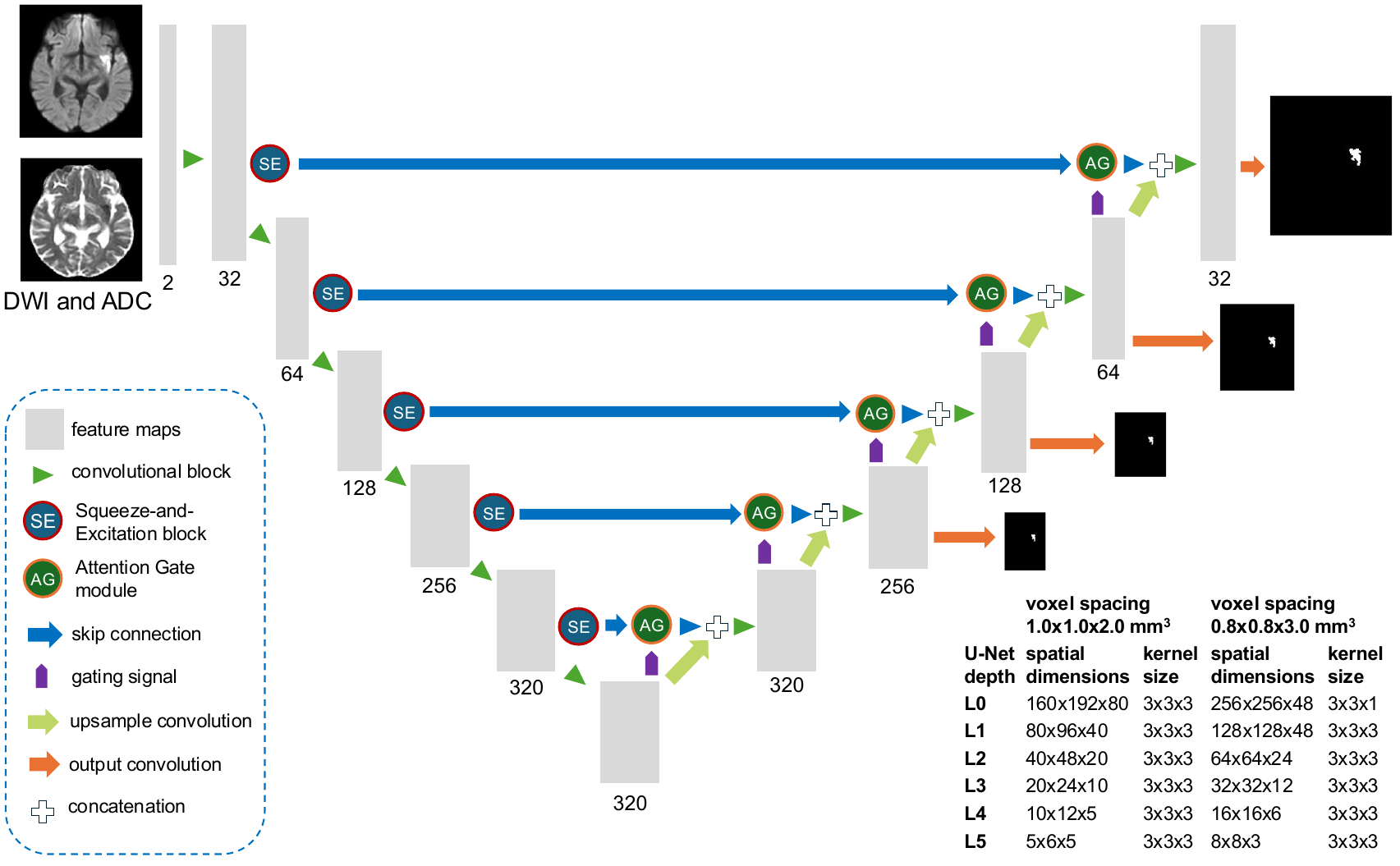}
    \caption{\textbf{Illustration of the U-Net architecture used in \textit{ISLA}.} The model takes a trace diffusion-weighted image (DWI) and the corresponding apparent diffusion coefficient map (ADC) as inputs. Lesion predictions are generated at multiple levels of the decoder for deep supervision. The number of feature maps at each level is indicated below the gray boxes. The lower-right table reports the spatial dimensions at each U-Net depth and the convolutional kernel sizes for the two input resolutions used in \textit{ISLA}.}
    \label{fig:isla_architecture}
\end{figure*}

The U-Net architecture used in \textit{ISLA} is represented in Figure \ref{fig:isla_architecture}. To exploit full volumetric information, we implemented a 3D U-Net that takes the whole-brain DWI and ADC volumes.

\subsubsection{Network architecture}
Several design choices were inspired by \textit{nnU-Net} \citep{Isensee2021}. In particular, the number of feature maps is initialized to 32 and doubled (halved) at each downsampling (upsampling) operation, with a maximum of 320 feature maps. For the highly anisotropic input images (0.8×0.8×3.0 mm$^3$), pooling is initially performed only in the in-plane dimensions, and two-dimensional convolutional kernels are used in the first encoder block and the last decoder block. Downsampling is implemented using strided convolutions, whereas upsampling is performed using transposed convolutions with kernel size equal to stride size.

\subsubsection{Deep supervision}
For deep supervision, intermediate outputs are generated at each U-Net level---except for the two deepest levels \citep{Isensee2021}---using 1×1×1 convolutions followed by sigmoid activation. During training, the ground-truth segmentation is downsampled using nearest-neighbor interpolation to match the spatial resolution of each auxiliary output. The overall loss is defined as
$$L = \sum_{i=1}^4 \alpha_i L_i$$
where $L_i$ denotes the loss between the ground truth segmentation and the $i$-th network output (with $i=1$ corresponding to the full-resolution output and $i=4$ to the lowest-resolution output), and $\alpha$ are the corresponding weights, defined as $\alpha = (0.53, 0.27, 0.13, 0.07)$.

\subsubsection{Attention mechanisms}
For channel attention, we used \textit{Squeeze-and-Excitation} (SE) block \citep{Hu2020}. Inspired by the \textit{Enc USE-Net} configuration of \citet{Rundo2019}, which demonstrated strong generalization performances across datasets, we placed an SE block after each encoder block to modulate the skip connections. Following \citeauthor{Rundo2019}, we set the reduction ratio to 8.

For spatial attention, we integrated \textit{Attention Gate} modules in the U-Net architecture \citep{Schlemper2019}. Specifically, a gating signal derived from coarse-scale feature maps is used to encode contextual information and modulate features propagated from the encoder (\ref{appendix_AG}).

\subsection{Ensemble approaches}
\label{meth:ensemble_approaches}
Starting from the baseline U-Net described above (Section \ref{meth:baseline_unet}), we constructed 8 variants that are combined at inference in \textit{ISLA} (Figure \ref{fig:isla_framework}).

\subsubsection{MRI resolutions}
The \textit{ISLA} ensemble strategy incorporates predictions obtained from two input resolutions: 1.0×1.0×2.0 mm$^3$ and 0.8×0.8×3.0 mm$^3$ (Section \ref{meth:preprocessing}). As shown in Figure \ref{fig:isla_architecture}, only isotropic kernel and stride sizes are used for the 1.0×1.0×2.0 mm$^3$ input images. In contrast, for the 0.8×0.8×3.0 mm$^3$ input images, the first encoder block and the last decoder block use two-dimensional convolutions, and downsampling is initially restricted to the in-plane dimensions.

\subsubsection{Convolutional blocks}
We implemented the convolutional blocks of the baseline U-Net architecture (Figure \ref{fig:isla_architecture}) in two configurations: standard convolutional blocks---StdUNet---and residual convolutional blocks---ResUNet.

For StdUNet, we adopted the convolutional blocks used in \textit{nnU-Net}. Each block consists of two sequences of a convolutional layer followed by Instance Normalization and a leaky ReLU activation (negative slope 0.01).

For ResUNet, we adapted the residual blocks proposed by \citet{Zhang2018} to 3D convolutions. Specifically, each block consists of two sequences of Instance Normalization, ReLU activation, and convolutional layer, except for the first block, which follows a \textit{Conv-InstNorm-ReLU-Conv} structure. Residual connections are formed by adding each block's input and output, with identity mappings implemented via 1×1×1 convolutions. We replaced the Batch Normalization used by \citeauthor{Zhang2018} with Instance Normalization, as the former is known to exhibit limited cross-domain generalization \citep{Li2018}.

In both the StdUNet and ResUNet encoders, downsampling is performed using strides in the first convolutional layer of each block. Furthermore, we added 3D Spatial Dropout---which promotes independence between feature maps to prevent overfitting \citep{Tompson2015}---in the three deepest encoder blocks. Specifically, it is inserted after the normalization layers, with a dropout rate of 0.2.

\subsubsection{Loss functions}
We evaluated two loss functions designed to address class imbalance in segmentation tasks: the sum of Dice loss \citep{Milletari2016} and binary cross-entropy (DiceBCE), and UFL.

For the Dice loss, a smoothing factor of 1 is used to stabilize training. For UFL, we adopted the asymmetric formulation and used the same hyperparameters as \citet{Yeung2022} for their 3D segmentation experiments: $\lambda = 0.5$, $\delta = 0.6$, and $\gamma = 0.5$.

\subsection{Ensemble inference}
\label{meth:ensemble_inference}
Based on the design variations included in \textit{ISLA} (Section \ref{meth:ensemble_approaches}), we implemented eight sub-models corresponding to all combinations of two MRI resolutions, two convolutional block types, and two loss functions. At inference time, test-time augmentation is additionally applied using left-right flipping. For each input resolution, the probability maps predicted by the corresponding four sub-models are averaged. The resulting probability maps are then transformed back to the original space by reversing the resampling and padding/cropping steps described in Section \ref{meth:preprocessing}, using linear interpolation for resampling. The probability maps corresponding to the two input resolutions are subsequently averaged, and a threshold of 0.5 is applied to obtain a binary segmentation (0: non-lesion, 1: lesion).

\subsection{\textit{ISLA-MT}: unsupervised domain adaptation}
\textit{ISLA-MT} is an extension of \textit{ISLA} that incorporates UDA to generalize better across MRI datasets. It is based on the MT framework proposed by \citet{Perone2019}.

In MT, the model comprises a student and a teacher, two neural networks sharing a common architecture. The student's weights are updated with stochastic gradient descent using the following loss function:
$$\mathcal{L} = \mathcal{L}_{sup} + \gamma \mathcal{L}_{cons}$$
where $\mathcal{L}_{sup}$ is the supervised loss, $\mathcal{L}_{cons}$ is a consistency loss penalizing the difference between student and teacher predictions for the same inputs, and $\gamma$ is the consistency weight hyperparameter. The teacher weights are updated using an exponential moving average (EMA):
$$\theta^\prime_t = \alpha \theta^\prime_{t-1} + (1-\alpha) \theta_t$$
where $\theta$ and $\theta^\prime$ denote the student and teacher weights, respectively, and $\alpha$ is the EMA decay hyperparameter.

\begin{figure*}
    \centering
    \includegraphics[scale=0.7]{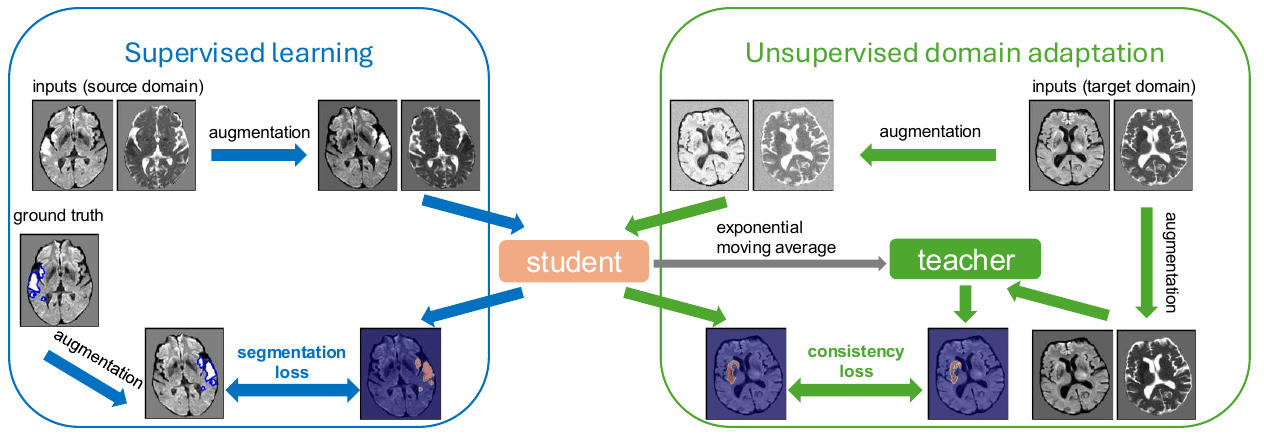}
    \caption{\textbf{Illustration of \textit{ISLA-MT} training.} Student training is guided by a weighted sum of segmentation and consistency losses. The segmentation loss is computed on labeled source-domain images, whereas the consistency loss is computed on unlabeled target-domain images. The teacher weights are updated using an exponential moving average of the student weights.}
    \label{fig:isla_mt_training}
\end{figure*}

In our MT implementation (Figure \ref{fig:isla_mt_training}), the consistency loss is computed as the mean squared difference between the student and teacher probability maps. As recommended by \citet{French2017} for UDA, this loss is calculated only on target-domain data. During a ramp-up phase, the EMA decay is set to 0.99 and then increased to 0.999 for the remainder of training \citep{Tarvainen2017}.

The consistency weight hyperparameter $\gamma$ is also gradually increased using a sigmoid ramp-up schedule \citep{Perone2019} during this ramp-up phase. Lower $\gamma$ values than those reported by \citeauthor{Perone2019} are used, as their study involved substantially fewer labeled training cases than ours (Section \ref{meth:data}). Accordingly, the final value of $\gamma$ at the end of the ramp-up phase was set to 5 when using the DiceBCE supervised loss and to 1 when using UFL, which typically produces loss values of lower magnitude.

For data augmentation in the target domain, identical spatial transformations are applied to both student and teacher inputs to ensure consistent supervision, while different intensity augmentations are applied to the student branch to improve robustness. Data augmentation details are provided in Section \ref{meth:training_config}.

In the teacher network, dropout is removed, as it has been reported to provide only marginal benefits \citep{Tarvainen2017}, as are the intermediate deep supervision outputs, which are thus only used during student supervised training. At inference time, predictions are generated using the teacher model.

\subsection{Training configuration}
\label{meth:training_config}
\textbf{Data augmentation.} The \textit{nnU-Net} data augmentation pipeline for 3D anisotropic images is integrated in \textit{ISLA}, with the addition of random translation and elastic deformations (\ref{appendix_dataAug}).

\textbf{Details.} Models are trained for 200 epochs with a batch size of 4 using stochastic gradient descent with Nesterov momentum of 0.99. The initial learning rate is set to 0.001 for DiceBCE and 0.005 for UFL, and is linearly decayed to $10^{-6}$ over training. Mixed precision training \citep{Micikevicius2017} is employed to reduce computational resource requirements. For \textit{ISLA-MT}, the batch size is set to 4 for both the supervised loss and the consistency loss, and the ramp-up phase lasts 40 epochs.

\section{Experimental setup}
\subsection{Data}
\label{meth:data}

\subsubsection{Training dataset}
Two public databases were used for training and internal validation of \textit{ISLA}: SOOP \citep{Absher2024} and ISLES 2022 \citep{HernandezPetzsche2022}. All participants had ischemic stroke and underwent MRI, including diffusion-weighted imaging and T2-weighted fluid-attenuated inversion recovery (FLAIR). MRI scans were acquired at various time points ranging from the acute to the subacute phase of stroke. Ischemic stroke lesions were manually delineated on DWI and ADC images.

\subsubsection{External cohort}
For an independent evaluation of segmentation performance, we included an external cohort of MR images acquired in routine clinical practice in the Hauts-de-France region.

This external cohort originates from a study classified as observational on March 9, 2010 by the \textit{Comité de protection des personnes Nord-Ouest IV}. The study was approved on December 21, 2010, by the committee responsible for protecting patient personal information (n° 10.677). The Data Protection Officer of Lille University Hospital certified that the project complied with applicable personal data protection regulations, including the General Data Protection Regulation (EU) 2016/679.

We included all patients treated by mechanical thrombectomy at Lille University Hospital between 2015 and 2022 for AIS due to large vessel occlusion diagnosed on brain MRI. All MRI examinations included at least diffusion-weighted imaging and FLAIR and were acquired at initial admission during the hyperacute to acute phases, before endovascular treatment.

The external cohort was divided into two subsets: the \textbf{external evaluation cohort}, comprising MRI examinations with manual AIS lesion annotations provided by three radiologists (MB, HD, and SY), and the \textbf{external adaptation cohort}, comprising unannotated MRI examinations. The latter was used for the unsupervised learning component of \textit{ISLA-MT} and included different patients who underwent MRI at the same hospital centers as those in the external evaluation cohort (20 centers).

Manual AIS lesion segmentation in the external evaluation cohort was performed on DWI using ITK-SNAP \citep{Yushkevich2006}. The readers, blinded to clinical data, independently delineated the lesions. They reviewed the images in the axial plane and traced all areas of hyperintense signal consistent with acute ischemia, slice by slice. In some cases, ADC and FLAIR images were also consulted to exclude artifacts and chronic lesions. Each case was annotated by one radiologist, while 28 cases were independently annotated by two radiologists to assess inter-rater agreement. For all analyses other than inter-rater agreement, only the annotations provided by the more experienced radiologist were used.

\subsubsection{Demographic and clinical information}
Table \ref{tab:datasets} summarizes the characteristics of the datasets. Except for ISLES, all DWI images had high in-plane resolution (approximately 1×1 mm$^2$) and relatively large slice thicknesses (greater than 3 mm). The external evaluation cohort contained larger lesions than the SOOP and ISLES datasets. Although Philips scanners predominated across all datasets, approximately half of the scans in the external cohort were acquired on Siemens or GE systems. Except for ISLES, most scans were acquired on 1.5 T scanners.

\begin{table*}
    \centering
    \caption{\textbf{Characteristics of populations, MRI scanners, diffusion-weighted imaging, and lesions for each dataset.}}
    \begin{tabular}{>{\raggedright}p{0.10\linewidth}>{\raggedright}p{0.12\linewidth}>{\raggedright}p{0.14\linewidth}>{\raggedright}p{0.14\linewidth}>{\raggedright}p{0.13\linewidth}>{\raggedright}p{0.10\linewidth}>{\raggedright\arraybackslash}p{0.11\linewidth}}
        \hline
        \textbf{dataset}&\textbf{number of participants}&\textbf{field strength, Tesla}&\textbf{manufacturer}&\textbf{in-plane pixel spacing, mm} $^a$&\textbf{slice thickness, mm} $^a$&\textbf{lesion size, mL} $^b$\\
        \hline
    
        SOOP&1265&\makecell[l]{1.5 (80.1\%)\\3.0 (19.9\%)}&\makecell[l]{Philips (98.3\%)\\GE (1.2\%)\\Siemens (0.5\%)}&0.95 $\pm$ 0.16&6.00 $\pm$ 0.04&7.98 (31.91)\\[15pt]

        ISLES&250&\makecell[l]{3.0 (57.6\%)\\unknown (42.4\%)}&\makecell[l]{Philips (50.0 \%)\\unknown (48.4\%)\\Siemens (1.6\%)}&1.88 $\pm$ 0.28&2.60 $\pm$ 1.15&6.66 (19.59)\\[15pt]

        \makecell[l]{external\\evaluation\\cohort}&590&\makecell[l]{1.5 (88.1\%)\\3.0 (11.9\%)}&\makecell[l]{Philips (46.6\%)\\Siemens (29.8\%)\\GE (23.6\%)}&1.01 $\pm$ 0.26&4.98 $\pm$ 0.85&15.28 (35.16)\\[15pt]

        \makecell[l]{external\\adaptation\\cohort}&1505&\makecell[l]{1.5 (91.5\%)\\3.0 (8.5\%)}&\makecell[l]{Philips (52.6\%)\\GE (28.3\%)\\Siemens (19.1\%)}&0.98 $\pm$ 0.22&5.12 $\pm$ 0.81&-\\[15pt]
        \hline
        \multicolumn{7}{l}{$^a$Values are expressed as mean $\pm$ standard deviation.}\\
         \multicolumn{7}{l}{$^b$Values are expressed as median (interquartile range).}\\
    \end{tabular}
    \label{tab:datasets}
\end{table*}

\begin{figure*}
    \begin{minipage}[c]{0.9\linewidth}
        \includegraphics{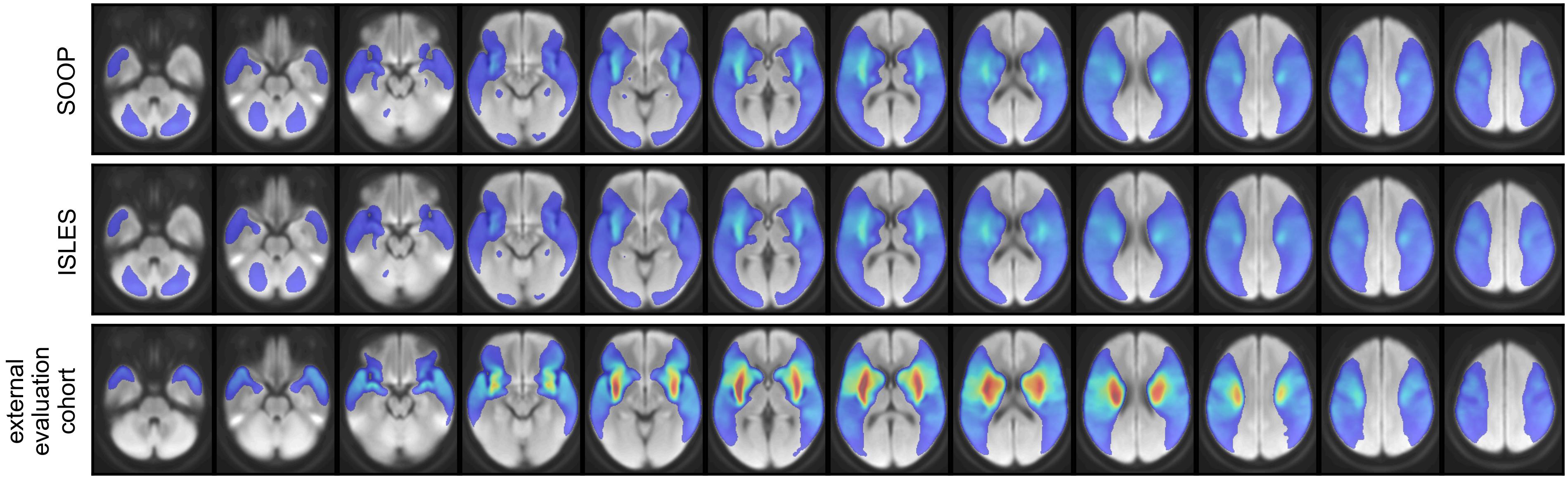}
    \end{minipage}
    \begin{minipage}[c]{0.06\linewidth}
        \includegraphics{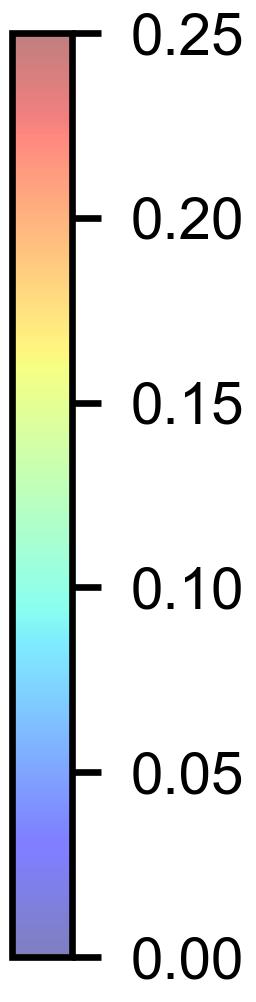}
    \end{minipage}
    \caption{\textbf{Proportions of ischemic stroke lesions across brain regions in the annotated datasets.} The manual lesion delineations were registered to the 1 mm$^3$ JHU ICBM template (Section \ref{meth:spatialError}), and voxel-wise proportions were computed. The proportion maps were then smoothed using a Gaussian filter (FWHM = 4 mm), and values below 0.02 were masked.}
    \label{fig:lesion_distrib}
\end{figure*}

As shown in Figure \ref{fig:lesion_distrib}, lesions were mostly distributed in the anterior circulation territories (internal carotid artery, middle cerebral artery, and anterior cerebral artery). In contrast, the SOOP and ISLES datasets showed more lesions in the posterior circulation territories.

\subsection{Comparison methods}
\label{meth:sota_methods}
We compared \textit{ISLA} and \textit{ISLA-MT} with three deep learning models developed for MRI-based AIS lesion segmentation: \textit{DAGMNet} \citep{Liu2021}, \textit{PAT} \citep{Jeong2024}, and \textit{DeepISLES} \citep{delaRosa2025}.

\textit{DAGMNet} and \textit{PAT} infer segmentation masks from DWI and ADC. For \textit{PAT}, we provided skull-stripped images generated using SynthStrip \citep{Hoopes2022}, with the brain mask computed from DWI, to match its training data; performance was lower without skull-stripping (data not shown). \textit{DeepISLES} combines the top-ranked models from the ISLES 2022 challenge. When FLAIR is available in addition to DWI and ADC, it combines the three top-ranked models; otherwise, it applies only the top-ranked model to DWI and ADC. The latter configuration was used for two cases of the external evaluation cohort for which FLAIR was unavailable. \textit{DAGMNet}, \textit{PAT}, and \textit{DeepISLES} were used as pre-trained models without additional training on our datasets\footnote{%
    \textit{DAGMNet}: \url{https://www.nitrc.org/projects/ads}
    accessed 2025-01-15.
    
    \textit{PAT}: \url{https://github.com/MDOpx/ISLES22-model-inference/tree/ae0846a}
    accessed 2026-05-12, variant DWI with transfer learning \& multi-tasking
    (the only working variant).
    
    \textit{DeepISLES}: \url{https://github.com/ezequieldlrosa/DeepIsles/tree/7658b60}
    accessed 2026-06-10.
}.

We additionally included \textit{nnU-Net} \citep{Isensee2021} in the comparison experiments, using the same training data as \textit{ISLA}. As \textit{nnU-Net} provides a complete end-to-end segmentation pipeline\footnote{\url{https://github.com/MIC-DKFZ/nnUNet/tree/v2.6.2} accessed 2026-01-03}, we applied minimal preprocessing to the DWI and ADC images, consisting of skull-stripping using SynthStrip, with the brain mask computed from DWI. Skull-stripping was necessary to ensure consistency within the training set, as the ISLES images used for training were already provided in skull-stripped form. We then used the full \textit{nnU-Net} pipeline, including automatic preprocessing, cross-validation-based selection of the best configuration, and ensemble inference.

\subsection{Segmentation metrics}
\label{meth:segmentation_metrics}
To quantify agreement between the predicted and ground truth segmentation masks, we used the metrics of the ISLES 2022 challenge: DSC and absolute volume difference (AVD, in mL) as voxel-level metrics, and absolute lesion count difference (ALD) and lesion detection F1 score (LF1) as lesion-level metrics \citep{HernandezPetzsche2022}. In addition, following previous AIS segmentation studies \citep{Clerigues2019,Liu2020,Platscher2022,Li2024}, we included the Hausdorff distance (HD) to account for distance (in mm) between the predicted and ground truth segmentation masks. Specifically, we computed the 95th percentile HD (HD95):
\begin{align*}
    &d(a,B) &&= \min_{b \in B} \|a-b\|_2
    \quad \text{point-to-set distance}\\
    &HD95(A,B) &&= \max \left\{
    \substack{
        Q_{0.95}\{d(a,B) : a \in A \}\\
        Q_{0.95}\{d(b,A) : b \in B \}
    }
    \right\}
\end{align*}

\subsection{Internal validation}
We conducted preliminary evaluations on the training dataset for internal validation. To this end, we held out 20\% of the SOOP images as a validation set (n=253) and used the remaining SOOP and ISLES images for training (n=1262). We did not include ISLES images in the validation set to enable comparisons with \textit{PAT} and \textit{DeepISLES}, which were trained using these images.

We first compared the \textit{ISLA} validation metrics with those obtained by its eight sub-models. We then compared the \textit{ISLA} validation metrics with those obtained using the comparison methods, applying two-tailed Wilcoxon signed-rank tests with multiple comparisons corrected using the Benjamini--Hochberg (BH) procedure.

\subsection{Generalization to an external cohort}
For application to the external evaluation cohort, \textit{nnU-Net}, \textit{ISLA} and \textit{ISLA-MT}  were trained using the full training dataset. The external adaptation cohort was additionally used as an unlabeled target domain for \textit{ISLA-MT}.

\subsubsection{Comparison in the overall population}
For each segmentation metric and for the four comparison methods, we conducted pairwise comparisons with \textit{ISLA} and \textit{ISLA-MT} using two-tailed Wilcoxon signed-rank tests. The same test was used to compare \textit{ISLA} and \textit{ISLA-MT} with each other. The resulting 45 p-values were corrected with the BH procedure.

Furthermore, for each model, we compared its segmentation metrics on the subset of 28 double-annotated cases with the corresponding inter-rater agreement metrics, using two-tailed Wilcoxon signed-rank tests with BH correction.

\subsubsection{Performance according to lesion characteristics}
Given the potential impact of lesion volume and conspicuity on segmentation quality, we analyzed these characteristics on the external evaluation cohort.

To quantify conspicuity, we computed a lesion contrast-to-noise ratio (CNR) for each case using the DWI, the ground truth lesion mask, and the SynthStrip-derived brain mask. To this end, a perilesional reference region is first defined by selecting all non-lesion brain voxels whose shortest distance to the lesion boundary lay between 2 and 10 mm. The CNR is then computed between the lesion and this perilesional reference region as:
$$
\mathrm{CNR} = \frac{\left|\mu_L - \mu_R\right|}{\sqrt{\sigma_L^2 + \sigma_R^2}},
$$
where $\mu_L$ and $\mu_R$ denote the mean intensities within the lesion and reference regions, respectively, and $\sigma_L^2$ and $\sigma_R^2$ denote the corresponding intensity variances. We intentionally included $\sigma_L^2$ as noise to account for lesion heterogeneity.

For each comparison method, we assessed the correlation between the DSC difference relative to \textit{ISLA} or \textit{ISLA-MT} and lesion volume and CNR, resulting in four correlations per comparison method. Specifically, Spearman's rank correlation coefficients and associated p-values were computed, and the 16 resulting p-values were corrected using the BH procedure.

\subsubsection{Spatial error analysis}
\label{meth:spatialError}
To visualize typical errors produced by the segmentation models, we computed false-positive (FP) and false-negative (FN) proportion maps. Specifically, for each case and each segmentation model, FP and FN maps were brought into a common space by nonlinearly registering the DWI images to the JHU ICBM-DTI-81 1 mm$^3$ template \citep{Mori2006} using ANTs, after skull-stripping with SynthStrip \citep{Hoopes2022}. The resulting transformations were then applied to the corresponding FP and FN maps, which were subsequently averaged across cases.

In addition to overall FP and FN maps aggregating all segmentation errors, we built maps aggregating only \textit{distant} model errors. We defined distant FPs (resp. FNs) as predicted (resp. ground-truth) lesion voxels located more than 5 mm from any ground-truth (resp. predicted) lesion voxel.

Based on the four error maps, we then selected representative cases illustrating typical model errors for visualization.

\subsubsection{Ablation studies}
\label{meth:ablations}
To assess the contribution of the different \textit{ISLA} components, we compared the full ensemble approach with reduced variants including only subsets of the sub-models. Specifically, given that \textit{ISLA} consists of a combination of two input resolutions, two types of convolutional blocks, and two loss functions (Section \ref{meth:ensemble_inference}), we evaluated ensemble variants in which one input resolution, one type of convolutional block, or one loss function was removed.

Then, for each variant, we compared each segmentation metric (Section \ref{meth:segmentation_metrics}) with \textit{ISLA} using two-tailed Wilcoxon signed-rank test---with p-values corrected with the BH procedure---and the associated rank-biserial correlation as an effect size measure \citep{Kerby2014}.

We conducted the same ablation experiment with \textit{ISLA-MT}.

\subsubsection{Computational efficiency}
For each evaluated model, we assessed the computational requirements of the complete inference pipeline on the entire external evaluation cohort, including loading, preprocessing, inference, postprocessing, and saving. Specifically, we measured inference time, peak GPU memory usage, and peak CPU RAM usage. We evaluated \textit{PAT}, \textit{DeepISLES}, \textit{nnU-Net}, \textit{ISLA}, and \textit{ISLA-MT} on a computer equipped with an Intel Xeon Gold 5418Y CPU (48 physical cores, 96 logical threads) and an NVIDIA L40S GPU. Because of compatibility and performance issues associated with \textit{DAGMNet}'s dependency on TensorFlow 2.0.0, released in 2019, we evaluated it on a separate machine equipped with an Intel Xeon W-2133 CPU (6 physical cores, 12 logical threads) and an NVIDIA Quadro RTX 6000 GPU.

\section{Results}

\subsection{Internal validation}
\begin{figure*}
    \centering
    \begin{minipage}[c]{0.8\linewidth}
        \centering
        \includegraphics{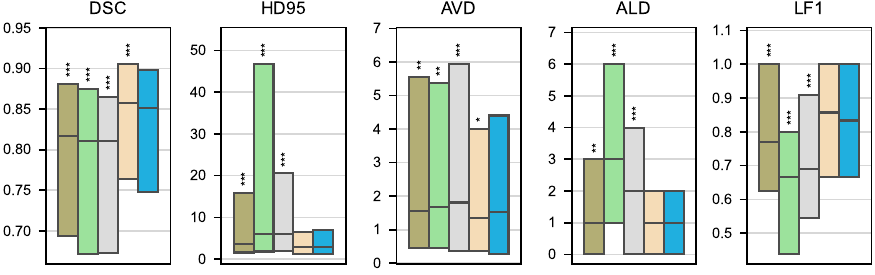}
    \end{minipage}
    \begin{minipage}[c]{0.17\linewidth}
        \centering
        \includegraphics{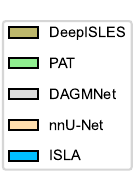}
    \end{minipage}
    \caption{\textbf{Segmentation metrics in the internal validation dataset.} Asterisks indicate significant pairwise comparisons ($^{***}$: $p < 0.001$, $^{**}$: $p<0.01$, $^*$: $p < 0.05$) with \textit{ISLA}. Boxplots display the median and interquartile range only; whiskers and outliers are omitted for clarity. Abbreviations $\rightarrow$ DSC: Dice similarity coefficient; HD95: 95th percentile Hausdorff distance; AVD: absolute volume difference; ALD: absolute lesion count difference; LF1: lesion detection F1 score.}
    \label{fig:internal_validation}
\end{figure*}

Overall, \textit{ISLA} outperformed its eight sub-models on the internal validation set across all segmentation metrics (\ref{appendix_val_isla_submodels}). Regarding the comparison methods, Figure \ref{fig:internal_validation} shows that the three pre-trained models (\textit{DeepISLES}, \textit{PAT}, and \textit{DAGMNet}) achieved lower performance than \textit{nnU-Net} and \textit{ISLA} across all metrics. Furthermore, \textit{nnU-Net} achieved significantly higher DSC and lower AVD than \textit{ISLA}.

\subsection{Generalization to an external cohort}

\subsubsection{Overall performance}
\label{res:metrics_global}
\begin{figure*}
    \centering
    \begin{minipage}[c]{0.8\linewidth}
        \centering
        \includegraphics{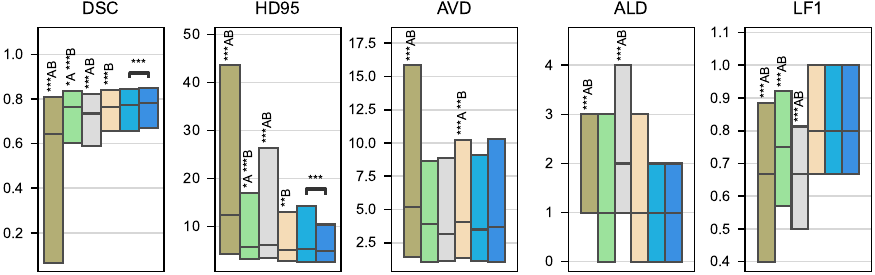}
    \end{minipage}
    \begin{minipage}[c]{0.17\linewidth}
        \centering
        \includegraphics{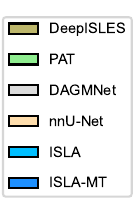}
    \end{minipage}
    \caption{\textbf{Segmentation metrics in the external evaluation cohort.} Asterisks indicate significant pairwise comparisons ($^{***}$: $p < 0.001$, $^{**}$: $p<0.01$, $^*$: $p < 0.05$), with a letter indicating the compared \textit{ISLA} model (A: \textit{ISLA}, B: \textit{ISLA-MT}). Boxplots display the median and interquartile range only; whiskers and outliers are omitted for clarity. Abbreviations $\rightarrow$ DSC: Dice similarity coefficient; HD95: 95th percentile Hausdorff distance; AVD: absolute volume difference; ALD: absolute lesion count difference; LF1: lesion detection F1 score.}
    \label{fig:metrics_global}
\end{figure*}

Figure \ref{fig:metrics_global} shows the performance of the evaluation methods across the segmentation metrics on the external evaluation cohort. None of the comparison methods outperformed \textit{ISLA} or \textit{ISLA-MT} on any metric. In contrast, both \textit{ISLA} methods significantly outperformed \textit{DeepISLES} across all metrics. The same trend was observed for \textit{DAGMNet}, except for AVD. Both proposed methods also significantly outperformed \textit{PAT} in DSC, HD95, and LF1. They also significantly outperformed \textit{nnU-Net} in AVD, while \textit{ISLA-MT} additionally significantly outperformed \textit{nnU-Net} in DSC and HD95. Regarding the comparison between the two proposed methods, \textit{ISLA-MT} significantly outperformed \textit{ISLA} across all metrics except ALD.

In the subset of 28 double-annotated cases, the inter-rater agreement metrics were significantly better than those obtained by all evaluated models (\ref{appendix_inter_rater}).

\subsubsection{Performance according to lesion characteristics}
\begin{figure*}
    \centering
    \begin{minipage}[c]{0.85\linewidth}
        \centering
        \includegraphics{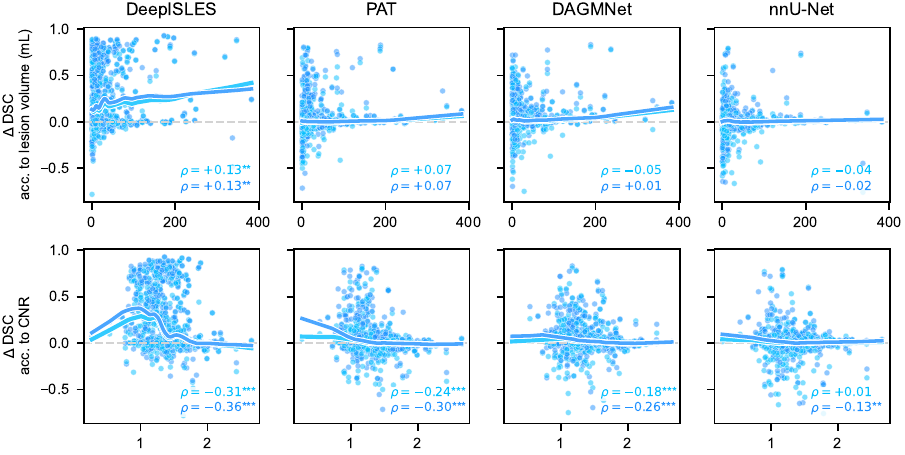}
    \end{minipage}
    \begin{minipage}[c]{0.13\linewidth}
        \centering
        \includegraphics{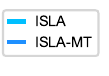}
    \end{minipage}
    \caption{\textbf{Difference in Dice similarity  coefficient ($\Delta$ DSC) according to lesion volume and contrast-to-noise ratio (CNR) in the external evaluation cohort.} For each comparison method and participant, the DSC difference relative to \textit{ISLA} or \textit{ISLA-MT} is reported ($\text{DSC}_{\textit{ISLA(-MT)}} - \text{DSC}_{\text{comparison}}$), along with the Spearman's rank correlation coefficient ($\rho$) and corresponding significance levels ($^{***}$: $p < 0.001$, $^{**}$: $p<0.01$, $^*$: $p < 0.05$). Curves show locally weighted linear regression fits of $\Delta$ DSC against lesion volume and CNR for each \textit{ISLA} and \textit{ISLA-MT} variant, computed with a span of 0.3 and 3 residual reweightings.}
    \label{fig:dsc_correlation}
\end{figure*}

Figure \ref{fig:dsc_correlation} shows the differences in DSC according to lesion volume and CNR. The correlation between lesion volume and DSC difference was significantly positive when comparing \textit{DeepISLES} with both \textit{ISLA} and \textit{ISLA-MT}; it was not significant for the other comparison methods. In contrast, the correlations between CNR and DSC difference were significantly negative in all cases except for the comparison between \textit{nnU-Net} and \textit{ISLA}. 

\subsubsection{Spatial error analysis}
\label{res:spatialErrors}
\begin{figure*}
    \centering
    \begin{minipage}[c]{0.05\linewidth}
        \rotatebox{90}{\textbf{overall}}
    \end{minipage}
    \begin{minipage}[c]{0.4\linewidth}
        \centering
        \textbf{false positives}\\
        \vspace{1em}
        \includegraphics{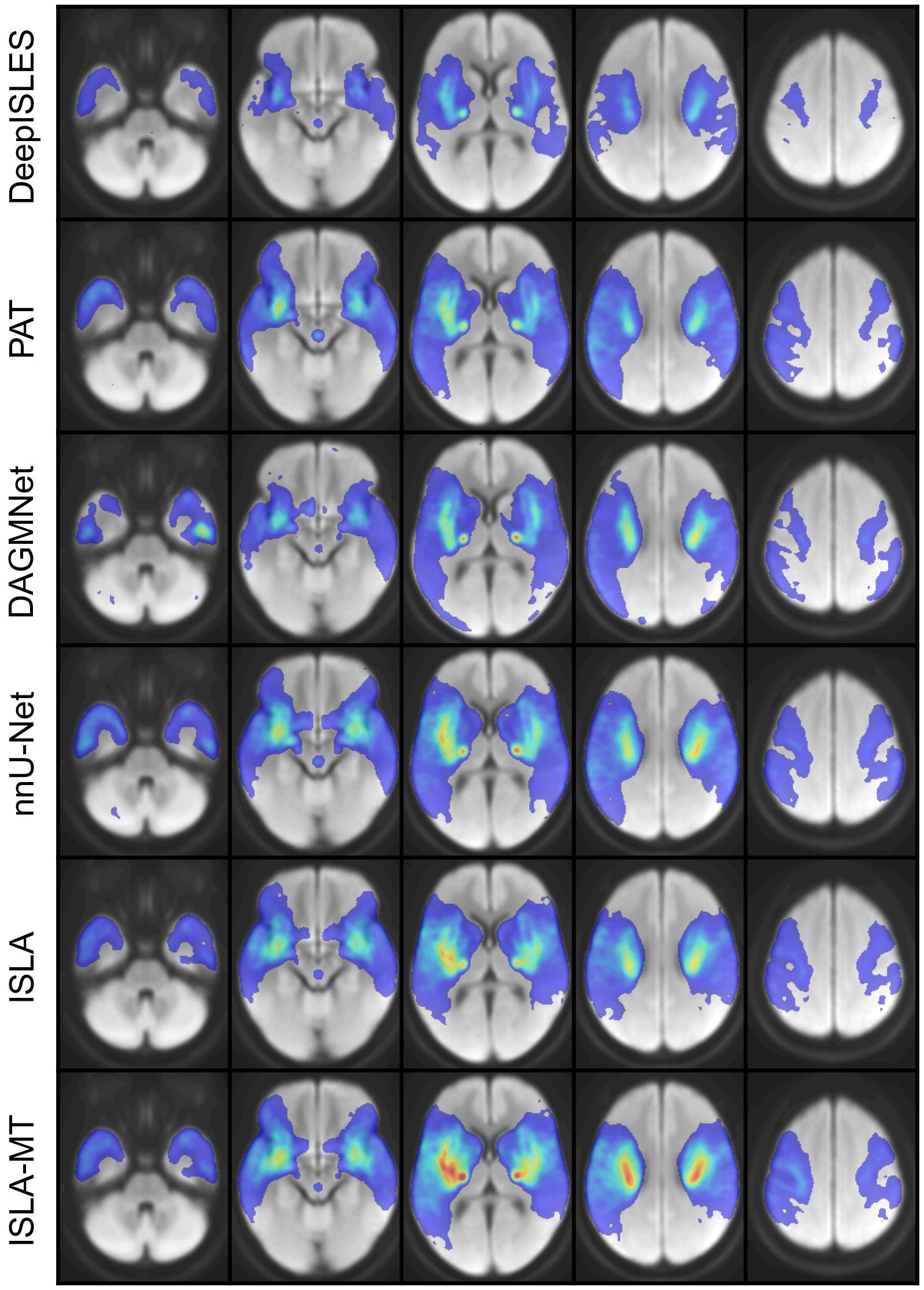}
    \end{minipage}
    \begin{minipage}{0.07\linewidth}
        \centering
        \includegraphics{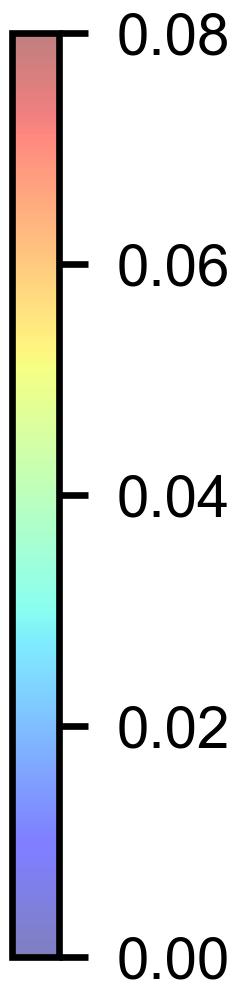}
    \end{minipage}
    \begin{minipage}[c]{0.38\linewidth}
        \centering
        \textbf{false negatives}\\
        \vspace{1em}
        \includegraphics{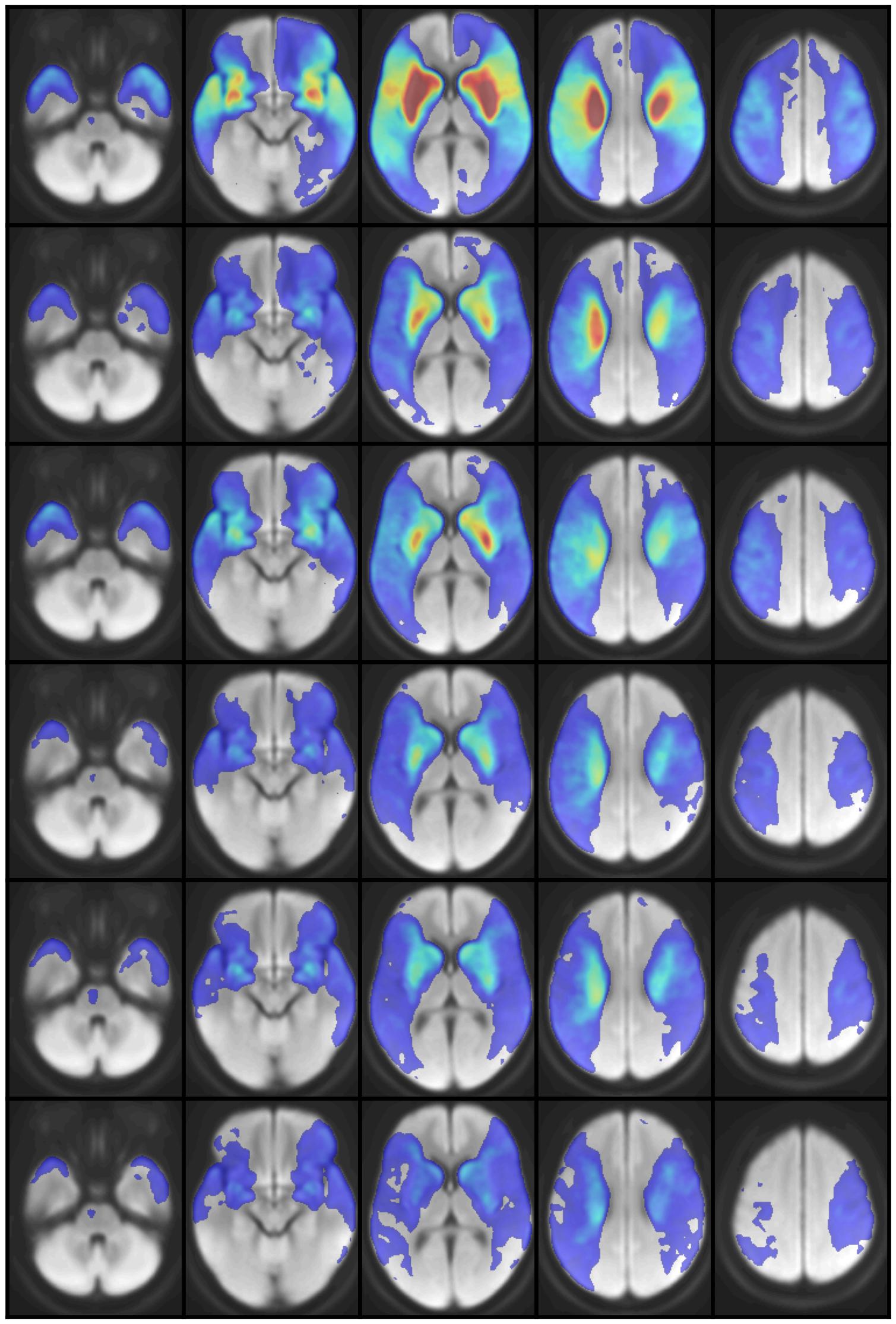}
    \end{minipage}

    \vspace{1em}
    \begin{minipage}[c]{0.05\linewidth}
        \rotatebox{90}{\textbf{distant}}
    \end{minipage}
    \begin{minipage}[c]{0.4\linewidth}
        \centering
        \includegraphics{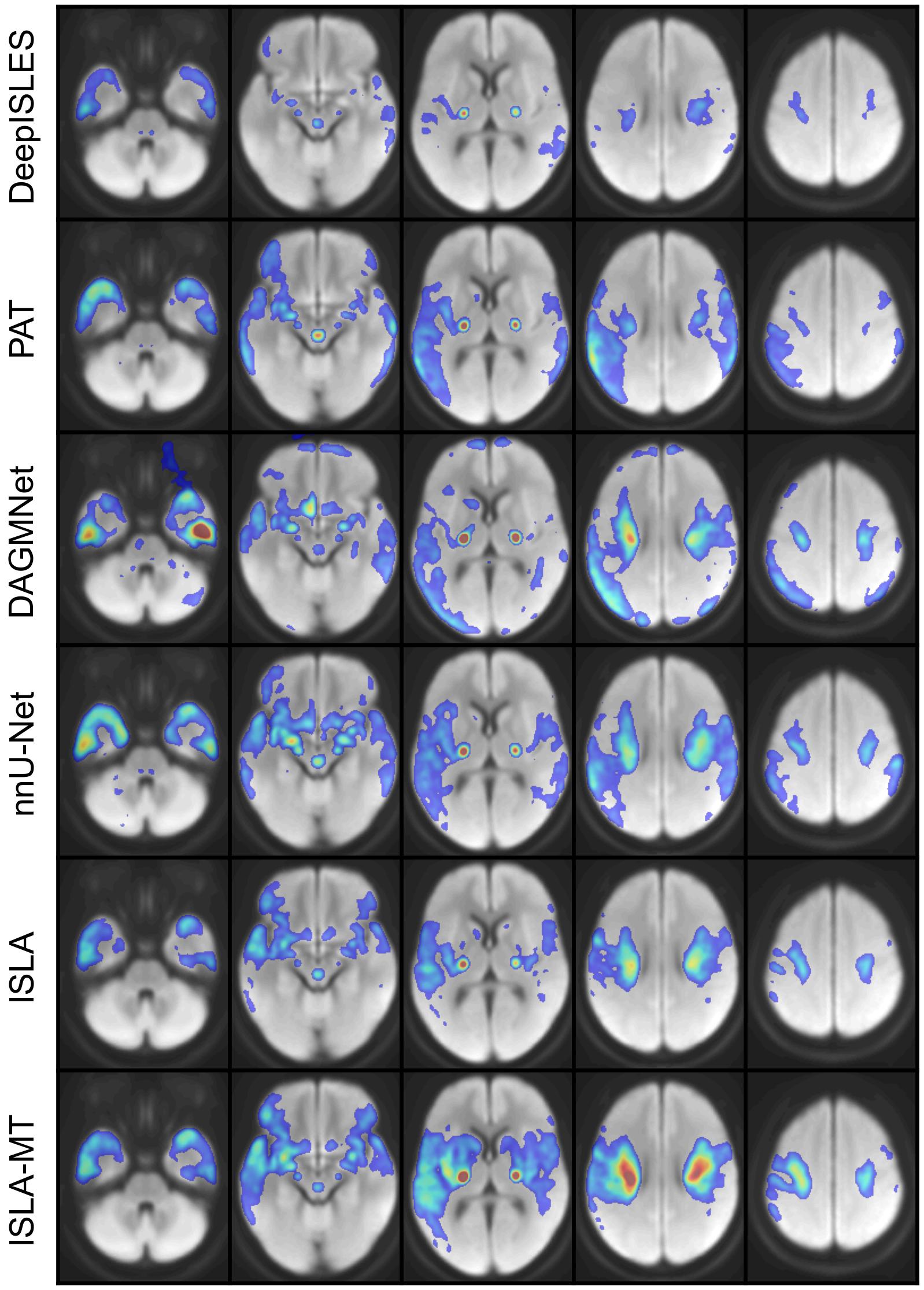}
    \end{minipage}
    \begin{minipage}{0.07\linewidth}
        \centering
        \includegraphics{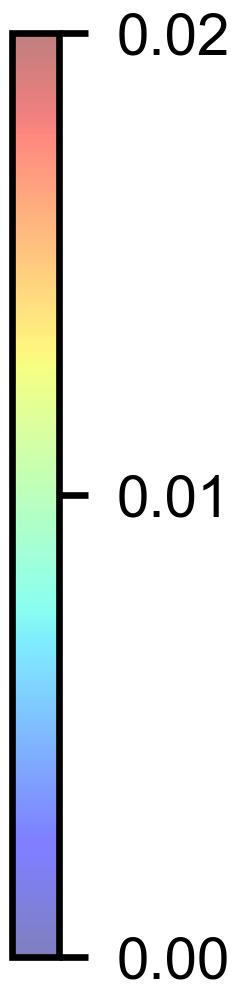}
    \end{minipage}
    \begin{minipage}[c]{0.38\linewidth}
        \centering
        \includegraphics{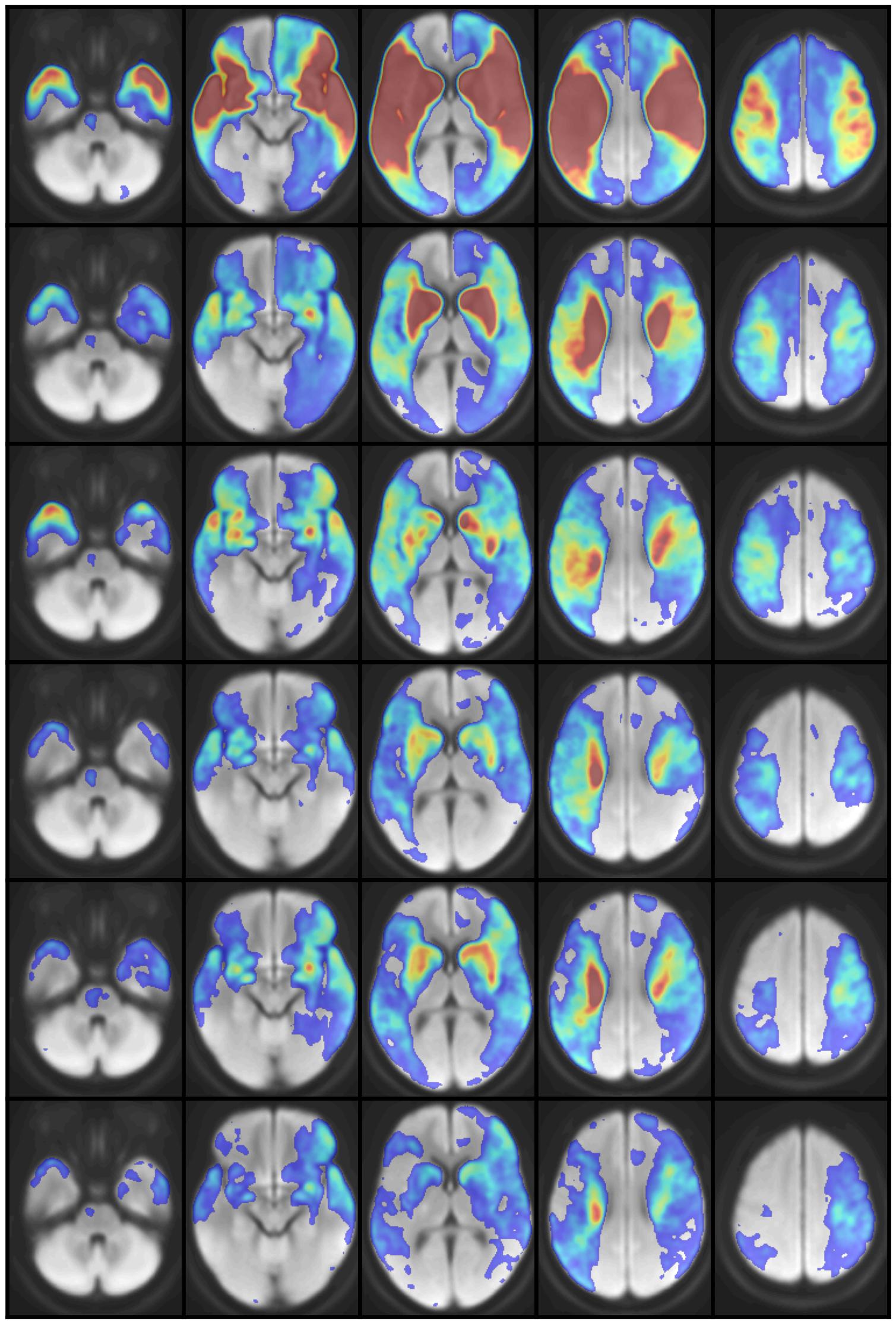}
    \end{minipage}
    \caption{\textbf{Voxel-wise proportions of false positives (FPs) and false negatives (FNs) in the external evaluation cohort.} Distant FPs (resp. FNs) correspond to predicted (resp. ground-truth) lesion voxels located more than 5 mm from any ground-truth (resp. predicted) lesion voxel. Error maps were transformed to MNI space before averaging. The average maps were smoothed using a Gaussian filter (FWHM = $4 \times 4 \times 8$ mm) and thresholded at 0.5\% for the overall error maps and 0.2\% for the distant error maps.}
    \label{fig:voxel_maps}
\end{figure*}

Several observations can be made from the voxel-wise maps of false positives (FPs) and false negatives (FNs) (Figure \ref{fig:voxel_maps}). First, \textit{DeepISLES} showed more extensive FNs than the other models across the brain, with relative sparing of the cerebellum and occipital lobe. Second, \textit{DAGMNet} generated substantial FPs in the temporal poles and anterior frontal regions. Third, \textit{DAGMNet} also generated substantial FPs anterior to the temporal poles, outside the brain. Fourth, \textit{PAT}, \textit{DAGMNet}, and \textit{nnU-Net} generated substantial FPs in the posterior superior part of the brain. Fifth, \textit{PAT} and \textit{DAGMNet} showed substantial proportions of FNs in the deep sylvian region. Sixth, although less pronounced than for \textit{DAGMNet} and \textit{PAT}, \textit{nnU-Net} and \textit{ISLA} also generated more FNs than \textit{ISLA-MT} in this region. Seventh, \textit{PAT}, \textit{DAGMNet}, and \textit{nnU-Net} produced more FNs than the \textit{ISLA} models in the left superior region of the brain. Eighth, \textit{ISLA-MT} generated slightly more FPs than the other models in the deep sylvian region.

\begin{figure*}
    \begin{minipage}{\linewidth}
        \centering
        \includegraphics{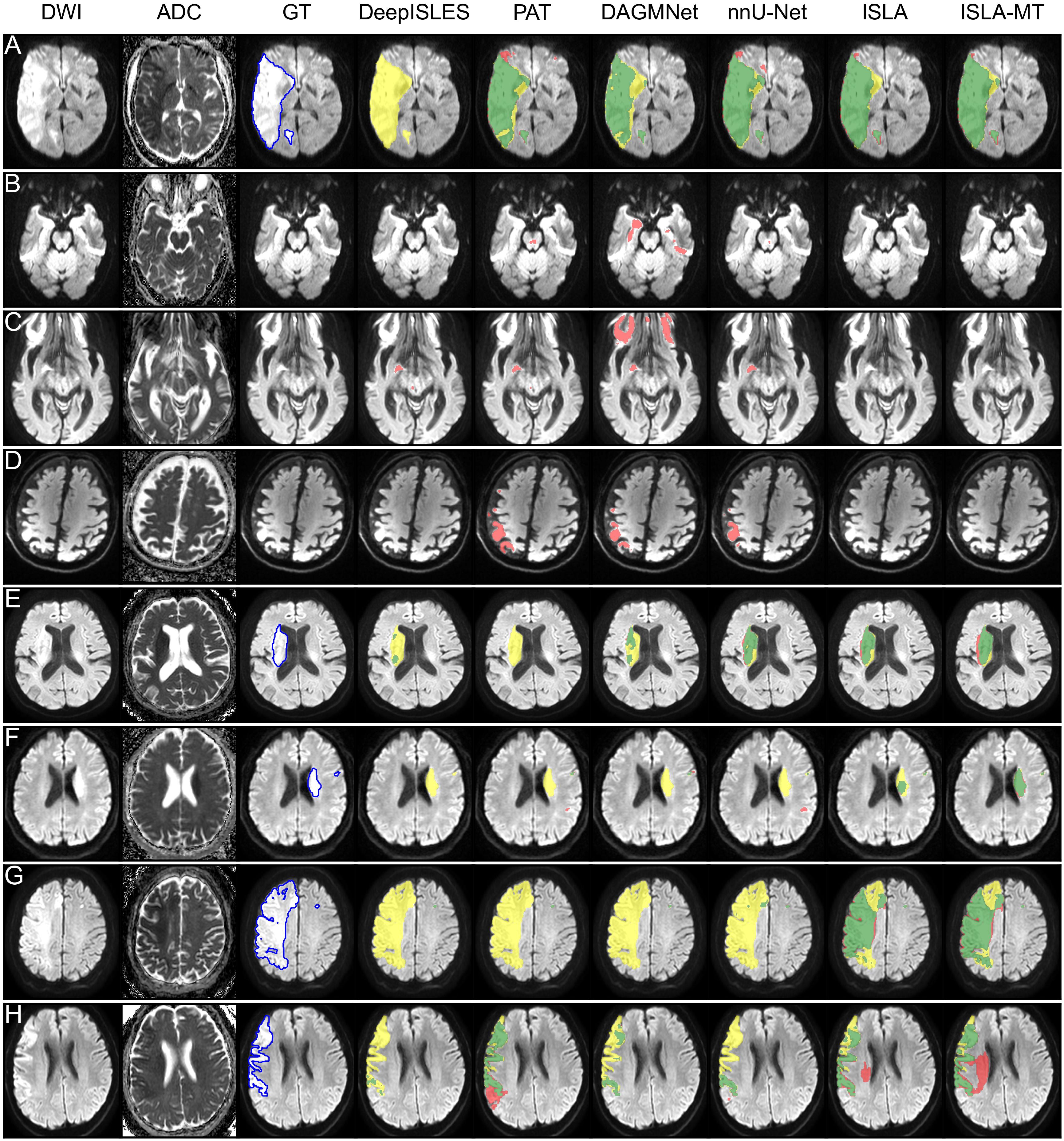}
    \end{minipage}
    \begin{minipage}{0.85\linewidth}
        \raggedleft
        \includegraphics{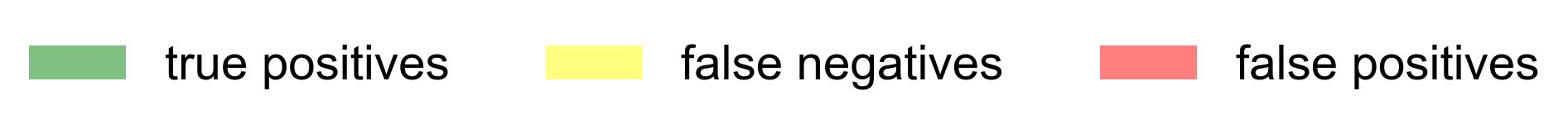}
    \end{minipage}
    \caption{\textbf{Illustrative examples of the different categories of model errors.} Axial slices of the trace diffusion-weighted image (DWI), apparent diffusion coefficient map (ADC), and ground-truth annotations (GT) are shown alongside true-positive, false-negative (FN), and false-positive (FP) voxels. All volumes were resampled to 0.8 × 0.8 × 3.0 mm$^3$ for visualization. \textbf{A} Large FN produced by \textit{DeepISLES}. \textbf{B} FP produced by \textit{DAGMNet} in the temporal poles and anterior frontal regions. \textbf{C} FP produced by \textit{DAGMNet} anterior to the temporal poles. \textbf{D} FP produced by \textit{PAT}/\textit{DAGMNet}/\textit{nnU-Net} in the posterior superior part of the brain. \textbf{E} FN produced by \textit{PAT}/\textit{DAGMNet} in the deep sylvian region. \textbf{F} FN produced by \textit{nnU-Net}/\textit{ISLA} in the deep sylvian region and detected by \textit{ISLA-MT}. \textbf{G} FN produced by \textit{PAT}/\textit{DAGMNet}/\textit{nnU-Net} in the left superior region of the brain. \textbf{H} FP produced by \textit{ISLA-MT} in the deep sylvian region.}
    \label{fig:cases_visualization}
\end{figure*}

Figure \ref{fig:cases_visualization} provides illustrative examples of each error category described above, with additional examples presented in \ref{appendix_illustrative_cases}. The FPs generated by \textit{DAGMNet} in the temporal poles were often caused by susceptibility artifacts near air-tissue interfaces. The FPs located anterior to the temporal poles were typically associated with distortion artifacts induced by magnetic field inhomogeneity. The FPs produced by \textit{PAT}, \textit{DAGMNet}, and \textit{nnU-Net} in the posterior superior part of the brain were attributable to low-spatial-frequency intensity inhomogeneity, which generated hyperintense regions in that area. The FNs observed in the deep sylvian region and the left superior part of the brain frequently coincided with low-contrast, heterogeneous lesions. Overall, \textit{ISLA} and \textit{ISLA-MT} appeared less sensitive to artifacts caused by motion and magnetic field inhomogeneity, particularly in brain regions near air-containing structures such as the frontal sinuses, petrous bones, and skull base. However, they appeared slightly more sensitive to motion-related anisotropy artifacts in the corticospinal tracts.

\subsubsection{Ablation studies}
\label{res:ablations}
The comparison results between the \textit{ISLA} models and their ablated variants are reported in Table \ref{tab:ablation_results}.

The input resolution had a limited effect on \textit{ISLA}, with the 1.0×1.0×2.0 mm$^3$ variant showing a significant decrease in DSC and increase in HD95, whereas no significant differences were observed for the 0.8×0.8×3.0 mm$^3$ variant. These effects were more pronounced with \textit{ISLA-MT}, for which both ablations significantly decreased DSC and increased HD95, with corresponding changes in AVD.

Regarding the model architecture, neither ablation significantly affected any metric for \textit{ISLA}. For \textit{ISLA-MT}, the ResUNet ablation resulted in a significant decrease in DSC and significant increases in HD95 and AVD.

The loss function ablations produced the largest changes: the DiceBCE ablations resulted in significant decreases in DSC and increases in AVD, whereas the opposite trends were observed for the UFL ablations. Furthermore, HD95 significantly increased for the DiceBCE ablation of \textit{ISLA-MT}.

Finally, although ALD increased and LF1 decreased for all ablations, these changes reached statistical significance only for a few variants.

\begin{table*}
    \centering
    \caption{\textbf{Comparison of segmentation metrics between the ablated and the full \textit{ISLA} variants in the external evaluation cohort.}}
    \begin{tabular}{>{\raggedright}p{0.25\linewidth}>{\raggedright}p{0.12\linewidth}>{\raggedright}p{0.12\linewidth}>{\raggedright}p{0.12\linewidth}>{\raggedright}p{0.12\linewidth}>{\raggedright\arraybackslash}p{0.12\linewidth}}
         \hline
         \textbf{ablated variant}&\textbf{DSC}&\textbf{HD95}&\textbf{AVD}&\textbf{ALD}&\textbf{LF1}\\[5pt]
         \hdashline
         &\multicolumn{5}{c}{\textbf{ISLA}}\\[2pt]
         
         \textbf{1.0×1.0×2.0mm$\bm{^3}$}&$-0.246^{***}$&$+0.131^{*}$&$-0.043^{}$&$+0.079^{}$&$-0.074^{}$\\[2pt]
         \textbf{0.8×0.8×3.0mm$\bm{^3}$}&$+0.037^{}$&$-0.026^{}$&$-0.109^{}$&$+0.006^{}$&$-0.101^{}$\\[4pt]
         
         \textbf{StdUNet}&$-0.056^{}$&$+0.067^{}$&$-0.025^{}$&$+0.063^{}$&$-0.097^{}$\\[2pt]
         \textbf{ResUNet}&$-0.015^{}$&$+0.051^{}$&$-0.112^{}$&$+0.059^{}$&$-0.052^{}$\\[4pt]
         
         \textbf{DiceBCE}&$-0.415^{***}$&$+0.063^{}$&$+0.395^{***}$&$+0.036^{}$&$-0.111^{}$\\[2pt]
         \textbf{UFL}&$+0.311^{***}$&$+0.044^{}$&$-0.357^{***}$&$+0.089^{}$&$-0.139^{*}$\\[5pt]
         \hdashline

         &\multicolumn{5}{c}{\textbf{ISLA-MT}}\\[2pt]

         \textbf{1.0×1.0×2.0mm$\bm{^3}$}&$-0.299^{***}$&$+0.301^{***}$&$-0.103^{*}$&$+0.108^{*}$&$-0.086^{}$\\[2pt]
         \textbf{0.8×0.8×3.0mm$\bm{^3}$}&$-0.195^{***}$&$+0.270^{***}$&$-0.128^{*}$&$+0.064^{}$&$-0.118^{*}$\\[4pt]
         
         \textbf{StdUNet}&$+0.000^{}$&$+0.143^{**}$&$-0.264^{***}$&$+0.061^{}$&$-0.071^{}$\\[2pt]
         \textbf{ResUNet}&$-0.540^{***}$&$+0.190^{***}$&$+0.584^{***}$&$+0.051^{}$&$-0.043^{}$\\[4pt]
         
         \textbf{DiceBCE}&$-0.631^{***}$&$+0.213^{***}$&$+0.615^{***}$&$+0.050^{}$&$-0.036^{}$\\[2pt]
         \textbf{UFL}&$+0.397^{***}$&$-0.022^{}$&$-0.494^{***}$&$+0.060^{}$&$-0.056^{}$\\[2pt]
         \hline
         \multicolumn{6}{p{0.95\linewidth}}{The name of each ablated variant indicates the component retained in the ensemble inference (e.g., only residual blocks in the ResUNet ablation). Rank-biserial correlation coefficients ($r_{rb}$) are reported, along with the corresponding significance levels ($^{***}$: $p < 0.001$, $^{**}$: $p < 0.01$, $^*$: $p < 0.05$). Negative and positive $r_{rb}$ values indicate a decrease and increase, respectively, in the corresponding metric due to ablation. Abbreviations $\rightarrow$ DSC: Dice similarity coefficient; HD95: 95th percentile Hausdorff distance; AVD: absolute volume difference; ALD: absolute lesion count difference; LF1: lesion detection F1 score.}
    \end{tabular}
    \label{tab:ablation_results}
\end{table*}

\subsubsection{Computational efficiency}
Table \ref{tab:comput_requir} reports the computational requirements for inference on the external evaluation cohort. \textit{ISLA} and \textit{ISLA-MT} were the fastest models, processing each case in around 5 seconds on average, closely followed by \textit{nnU-Net} and \textit{PAT}. Processing time was around 20 times longer for \textit{DeepISLES} and 30 times longer for \textit{DAGMNet}. Peak GPU memory usage ranged from 8 to 10 GiB for \textit{PAT}, \textit{DAGMNet}, \textit{ISLA}, and \textit{ISLA-MT}, compared with substantially higher usage for \textit{DeepISLES} and lower usage for \textit{nnU-Net}. Peak CPU RAM usage exceeded 20 GiB for \textit{ISLA} and \textit{ISLA-MT}, substantially higher than for the other models, particularly \textit{PAT} and \textit{DAGMNet}, which required less than 5 GiB.

\begin{table*}
    \centering
    \caption{\textbf{Computational requirements for inference on the external evaluation cohort.}}
    \begin{tabular}{>{\raggedright}p{0.16\linewidth}>{\raggedright}p{0.16\linewidth}>{\raggedright}p{0.16\linewidth}>{\raggedright}p{0.18\linewidth}>{\raggedright\arraybackslash}p{0.18\linewidth}}
        \hline
        \multirow{2}{*}{}&\multicolumn{2}{c}{\textbf{runtime}}&\multirow{2}{\linewidth}{\textbf{peak GPU memory (GiB)}}&\multirow{2}{\linewidth}{\textbf{peak CPU RAM (GiB)}}\\
        &\textbf{total}&\textbf{average / case}&&\\[4pt]
        \hline
        \textbf{DeepISLES}&18 h 11 min 39 s&1 min 51 s&19.80&9.54\\[2pt]
        \textbf{PAT}&57 min 24 s&6 s&8.87&3.55\\[2pt]
        \textbf{DAGMNet}&27 h 9 min 31 s&2 min 46 s&8.62&4.62\\[2pt]
        \textbf{nnU-Net}&52 min 11 s&5 s&3.77&8.87\\[2pt]
        \textbf{ISLA}&51 min 58 s&5 s&8.48&22.62\\[2pt]
        \textbf{ISLA-MT}&51 min 47 s&5 s&8.48&22.48\\[2pt]
        \hline
        \multicolumn{5}{p{0.95\linewidth}}{Computational requirements were measured for the complete inference pipeline on the external evaluation cohort, including loading, preprocessing, model inference, postprocessing, and saving the results.}
    \end{tabular}
    \label{tab:comput_requir}
\end{table*}

\section{Discussion}
We presented \textit{ISLA}, a publicly available deep learning framework for AIS lesion segmentation from brain diffusion MRI. \textit{ISLA} builds upon the successful medical image segmentation framework \textit{nnU-Net}, notably by integrating attention mechanisms and an ensemble-learning strategy that combines different input resolutions, convolutional block types, and loss functions. Furthermore, \textit{ISLA-MT}, a semi-supervised variant of \textit{ISLA}, is designed to improve generalization to external datasets.

We evaluated \textit{ISLA} on an internal validation set of 253 cases and both \textit{ISLA} variants on an external multicenter clinical cohort of 590 cases. \textit{ISLA} outperformed three pre-trained AIS lesion segmentation methods on the internal validation set, while both \textit{ISLA} and \textit{ISLA-MT} outperformed these methods on the external cohort, with consistent results across voxel-level, lesion-level, and distance-based segmentation metrics. Although \textit{nnU-Net} performed slightly better than \textit{ISLA} on the internal validation set, the external evaluation demonstrated the stronger generalization of the \textit{ISLA} models, particularly in detecting low-conspicuity lesions. \textit{ISLA-MT} further improved performance on the external cohort. Spatial error analyses and illustrative examples also suggested that the \textit{ISLA} models were generally more robust to acquisition artifacts commonly encountered in clinical practice.

\subsection{Comparison against pre-trained state-of-the-art methods}
Among the three pre-trained methods for AIS lesion segmentation evaluated in this study, \textit{DeepISLES} performed the worst on the external evaluation cohort. Voxel-wise FN maps and the associated illustrative examples showed that \textit{DeepISLES} tended to be conservative, frequently missing lesions. This observation was further supported by supplementary analyses showing substantially lower recall for \textit{DeepISLES} than for the other models (\ref{appendix_supplementaryMetrics}). As \textit{DeepISLES} is based on majority voting across three sub-models, we additionally evaluated each constituent model and observed similarly low performance for all three compared to the \textit{ISLA} approaches (\ref{appendix_metric_deepISLES_subModels}).

Although \textit{PAT} and \textit{DAGMNet} performed slightly worse than \textit{DeepISLES} on the internal validation set, the performance gap between these methods and the \textit{ISLA} models was narrower on the external evaluation cohort. Nevertheless, the \textit{ISLA} models significantly outperformed both methods across all metrics, except for AVD and for ALD when comparing with \textit{PAT}. The poor performance of \textit{DAGMNet} was particularly pronounced for HD95 and the lesion-level metrics (ALD and LF1). Joint inspection of the spatial error analyses and illustrative examples suggests that this may be partly explained by \textit{DAGMNet}'s sensitivity to MRI artifacts, which appeared to lead to a high number of small FPs.

\subsection{Comparison against \textit{nnU-Net}}
\textit{nnU-Net} achieved segmentation performance closer to that of the \textit{ISLA} models than the pre-trained approaches. Specifically, \textit{nnU-Net} slightly outperformed \textit{ISLA} in DSC and AVD on the internal validation set. On the external evaluation cohort, however, \textit{ISLA} and \textit{ISLA-MT} outperformed \textit{nnU-Net} in AVD, with \textit{ISLA-MT} also achieving better performance in DSC and HD95.

This discrepancy between in-domain and out-of-domain performance may partly reflect differences in the underlying approaches. \textit{nnU-Net} is a self-configuring pipeline that adapts to a dataset by selecting network architectures, postprocessing, and ensemble strategies based on internal cross-validation \citep{Isensee2021}. By contrast, our framework combines models with different technical characteristics through ensemble learning, without selecting a single best configuration, with the aim of improving generalization. For example, the comparison of \textit{ISLA}'s sub-models on the internal validation set favored the \textit{DiceBCE} loss, whereas the ablation study on the external evaluation cohort favored UFL. The use of spatial dropout during training and bias field inhomogeneity correction may also contribute to the improved generalization of the \textit{ISLA} models. Previous MRI studies have reported performance degradation and overconfidence in \textit{nnU-Net} when applied to out-of-domain data \citep{Vorberg2025,Hong2026}.

\subsection{Influence of lesion conspicuity}
Because ischemic lesions become hyperintense on DWI within the first minutes to hours after stroke onset, their signal intensity during the hyperacute phase can be heterogeneous, ranging from marked hyperintensity to more subtle changes. This results in variable lesion conspicuity. The analysis of DSC differences according to CNR showed that the performance advantage of the \textit{ISLA} models over the comparison methods increased as lesion conspicuity decreased, particularly for \textit{ISLA-MT}. Although lesion conspicuity may be influenced by multiple factors, these results suggest that the \textit{ISLA} models may be particularly well suited to segmenting lesions in acute and hyperacute ischemic stroke, when subtle diffusion abnormalities can make lesion delineation challenging \citep{Szocs2026}. The predominantly hyperacute-to-acute nature of the lesions in the external evaluation cohort may also partly explain the relatively poor performance of \textit{DeepISLES} and \textit{DAGMNet}, as their original publications report reduced performance during these phases \citep{Liu2021,delaRosa2025}. Stroke phase may likewise contribute to the limited performance of \textit{PAT}, which, like \textit{DeepISLES}, was trained on the ISLES 2022 dataset. Of its 250 images, 198 originated from Center \#1, whose cases ranged from the acute to early subacute phases \citep{HernandezPetzsche2022}. These findings are clinically relevant, as early identification and assessment of the core infarct are central to international recommendations for AIS management \citep{Prabhakaran2026}.

\subsection{Unsupervised domain adaptation}
The UDA approach implemented in \textit{ISLA-MT} yielded significant improvements over \textit{ISLA} in DSC and HD95 on the external evaluation cohort. The analysis of DSC differences according to CNR suggests that these improvements were primarily driven by better performance on low-conspicuity lesions.

Similar improvements in model generalization have been reported by \citet{Perone2019}, who used a related MT approach for spinal cord gray matter segmentation, and by \citet{Wang2023}, who employed a source-free domain adaptation strategy for AIS lesion segmentation.

Moreover, the \textit{ISLA} framework incorporates an exhaustive data augmentation pipeline that extends the \textit{nnU-Net} pipeline, which may further reduce the benefits of domain adaptation \citep{Campello2021}. Nonetheless, although the improvements were modest, they were statistically significant and suggest that \textit{ISLA-MT} may provide additional benefits on external datasets exhibiting larger domain shifts.

\subsection{Ensemble learning}
\textit{ISLA} leverages multiple variants of a baseline U-Net architecture for ensemble inference. Although additional design variations could have been incorporated into \textit{ISLA}---such as the presence or absence of deep supervision or attention mechanisms---we focused on three factors for which no clear consensus exists in the literature and that represent important sources of model variability: input resolution, convolutional block type, and loss function. Previous studies have highlighted the benefits of introducing algorithmic diversity in ensemble learning for brain MRI segmentation \citep{Li2022,Wiltgen2024,delaRosa2025}.

However, not all ablation experiments resulted in significant performance declines, particularly those involving \textit{ISLA}. Notably, the UFL ablations significantly improved DSC and AVD. Additional analyses revealed that almost all ablations significantly improved precision at the expense of recall, or vice versa (\ref{appendix_ablation_precision_recall}). Consequently, when the goal is to achieve robust generalization across datasets, including all submodels may remain the best option. However, the ensemble could be adapted to a specific dataset depending on whether FNs or FPs are more critical or challenging to address.

\subsection{Reproducibility and computational efficiency}
Limited reproducibility of deep learning tools has been identified as an important challenge in medical imaging \citep{Simko2024,Bontempi2024}, including MRI-based AIS lesion segmentation \citep{Baaklini2025}. In this study, we promote reproducibility by providing a public repository containing the complete processing pipeline for both training and inference, along with pre-trained model weights.

Furthermore, we assessed the computational requirements of the evaluated models to investigate their suitability for clinical workflows. \textit{ISLA} and \textit{ISLA-MT} achieved processing times of approximately 5 seconds per case, comparable to those of \textit{nnU-Net} and \textit{PAT}, and substantially shorter than those of \textit{DeepISLES} and \textit{DAGMNet}. Despite comprising eight submodels, \textit{ISLA} and \textit{ISLA-MT} maintained a moderate peak GPU memory usage of 8.48 GiB, compatible with a wide range of modern GPUs. Although their peak CPU RAM usage was higher than that of the other models, at approximately 22.5 GiB, this requirement remains within the memory capacity of most modern workstations.

\subsection{Limitations and perspectives}
Each ground truth segmentation in the external evaluation cohort was produced by a single radiologist. Although this approach is common in large-scale annotated datasets \citep{Absher2024}, and inter-rater agreement on the double-annotated cases was higher than the performance of all segmentation models, ground truth annotations could be improved through review by multiple experts \citep{HernandezPetzsche2022,Liu2023}.

Despite the growing number of models incorporating Transformer-based architectures for stroke lesion segmentation \citep{ZafariGhadim2024}, we restricted \textit{ISLA} to U-Net-based architectures. This choice was motivated by the success of \textit{nnU-Net} in medical image segmentation \citep{Campello2021,Antonelli2022} and by its predominance in AIS lesion segmentation \citep{Abbasi2023,Luo2024,Baaklini2025}. Nevertheless, combining U-Net and Transformer-based models may further increase algorithmic diversity, which could be beneficial for ensemble-based inference.

Another limitation of our study concerns the UDA setting. Although \textit{ISLA-MT} demonstrated significant improvements over \textit{ISLA}, the dataset used for unsupervised learning was substantial (1505 participants). Future studies could investigate the performance of \textit{ISLA-MT} when fewer target-domain MR images are available, thereby assessing its robustness under more limited data availability.

\section{Conclusion}
We introduce \textit{ISLA}, a novel ensemble-based deep learning framework for AIS lesion segmentation from diffusion MRI. \textit{ISLA} combines attention mechanisms, deep supervision, and ensemble learning across multiple input resolutions, convolutional block designs, and loss functions. We further propose \textit{ISLA-MT}, an extension incorporating UDA through the MT framework to improve generalization to external datasets. \textit{ISLA} significantly outperformed three pre-trained AIS lesion segmentation models on both an internal validation set and an external cohort. Although \textit{ISLA} was slightly outperformed by a \textit{nnU-Net} model trained on the same data during internal validation, it demonstrated better generalization to the external cohort. \textit{ISLA-MT} further improved segmentation performance on the external cohort, highlighting the potential of leveraging unlabeled target-domain data. These performance gains were particularly evident for low-conspicuity lesions and in the presence of acquisition-related artifacts. The proposed models, along with their training and inference pipelines and pre-trained weights, are publicly available to facilitate reproducibility and further research in stroke lesion segmentation.

\appendix

\section{Details of the Attention Gate modules in \textit{ISLA}}
\label{appendix_AG}
\setcounter{figure}{0}
Figure \ref{fig:attention_gate} illustrates our implementation of the Attention Gate module.

\begin{figure*}
    \centering
    \includegraphics[scale=0.6]{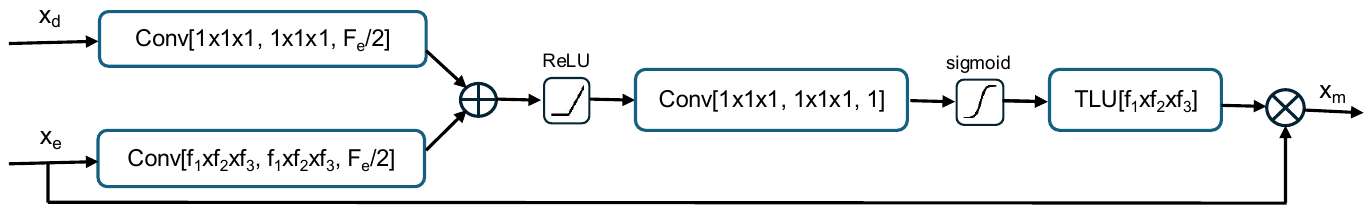}
    \caption{\textbf{Illustration of the implemented attention gate module.} $x_d$ denotes the decoder feature maps, $x_e$ denotes the encoder feature maps, and $x_m$ the output modulated feature maps. $Conv[k_1 \times k_2 \times k_3, s_1 \times s_2 \times s_3, F]$ represents a convolution with $F$ filters, kernel size $k_1 \times k_2 \times k_3$, and stride $s_1 \times s_2 \times s_3$. $TLU[u_1 \times u_2 \times u_3]$ denotes trilinear upsampling with factors $u_1 \times u_2 \times u_3$.}
    \label{fig:attention_gate}
\end{figure*}

\section{Details of \textit{ISLA}'s data augmentation}
\label{appendix_dataAug}
During \textit{ISLA} training, images are randomly augmented \textit{on the fly} to increase variability. These transformations are computed using background CPU workers. The following augmentations are applied (in the order listed below):
\begin{enumerate}
    \item \textbf{Flipping.} Left-right flipping is applied with a probability of 0.5.
    \item \textbf{Rotation.} Rotation around the superior-inferior axis is applied with a probability of 0.2, with the rotation angle sampled from $\mathcal{U}(-30,30)$. Third-order spline interpolation is used.
    \item \textbf{Scaling.} Scaling in the axial plane is applied with a probability of 0.2, with the scaling factor sampled from $\mathcal{U}(0.7,1.4)$. Linear interpolation is used.
    \item \textbf{Elastic deformation.} Elastic deformation \citep{vanTulder2024} is applied in the axial plane with a probability of 0.2. For the 1.0×1.0×2.0 mm$^3$ input images, 3×3 control points are used, and the displacement values in the left-right and anterior-posterior axes are sampled from $\mathcal{N}(0,4^2)$. For the 0.8×0.8×3.0 mm$^3$ input images, 4×4 control points are used, and the displacement values in the left-right and anterior-posterior axes are sampled from $\mathcal{N}(0,5^2)$. Third-order spline interpolation is used.
    \item \textbf{Translation.} Translation is applied with a probability of 0.5. Displacements are sampled from $\mathcal{U}(-20,20)$ voxels for the left-right and anterior-posterior axes, and from $\mathcal{U}(-10,10)$ for the superior-inferior axis.
    \item \textbf{Gaussian noise.} Gaussian noise is added to all voxels with a probability of 0.15. The noise standard deviation $\sigma$ is sampled from $\mathcal{U}(0,0.3)$, and each voxel noise is sampled from $\mathcal{N}(0,\sigma^2)$.
    \item \textbf{Gaussian blur.} Gaussian blur is applied in the axial plane with a probability of 0.2 per sample. When triggered, it is applied with a probability of 0.5 independently for each input image (DWI and ADC). The standard deviation of the gaussian kernel is sampled from $\mathcal{U}(0.5,1.5)$ for each input image.
    \item \textbf{Brightness update.} With a probability of 0.15, voxel intensities are multiplied by a value sampled from $\mathcal{U}(0.7,1.3)$.
    \item \textbf{Contrast.} With a probability of 0.15, voxel intensities are multiplied by a value sampled from $\mathcal{U}(0.7,1.3)$. Afterward, each input image is clipped to its original value range.
    \item \textbf{Simulation of low resolution.} This transformation is applied in the axial plane with a probability of 0.25. When triggered, it is applied with a probability of 0.5 independently for each input image. The image is downsampled by a factor sampled from $\mathcal{U}(1,2)$ using nearest-neighbor interpolation and then upsampled back to its original dimension using third-order spline interpolation.
    \item \textbf{Gamma augmentation.} With a probability of 0.15, gamma augmentation is applied to the input images, with a gamma factor sampled from $\mathcal{U}(0.7,1.5)$. With a probability of 0.15, intensity inversion is applied before gamma correction: $i_{new} = 1 - (1-i_{old})^\gamma$.
\end{enumerate}

\section{Comparison of \textit{ISLA} and its sub-models on the internal validation set}
\label{appendix_val_isla_submodels}
\setcounter{table}{0}
The segmentation metrics of \textit{ISLA} and its 8 sub-models on the internal validation set are shown in Table \ref{tab:val_isla_submodels}.

\begin{table*}
    \centering
    \caption{\textbf{Comparison of \textit{ISLA} and its sub-models on the internal validation set.}}
    \begin{tabular}{>{\raggedright}p{0.25\linewidth}>{\raggedright}p{0.12\linewidth}>{\raggedright}p{0.12\linewidth}>{\raggedright}p{0.12\linewidth}>{\raggedright}p{0.09\linewidth}>{\raggedright\arraybackslash}p{0.12\linewidth}}
        \hline
        &\textbf{DSC}&\textbf{HD95}&\textbf{AVD}&\textbf{ALD}&\textbf{LF1}\\[3pt]
        \textbf{ISLA}&0.852 (0.150)&2.950 (5.950)&1.521 (4.145)&1 (2)&0.833 (0.333)\\[2pt]
        \textbf{ResUNet-DiceBCE-0.8}&0.850 (0.144)&3.067 (6.611)&1.379 (3.705)&1 (2)&0.842 (0.333)\\[2pt]
        \textbf{ResUNet-UFL-0.8}&0.848 (0.161)&3.088 (6.031)&1.463 (4.322)&1 (2)&0.800 (0.333)\\[2pt]
        \textbf{StdUNet-DiceBCE-1.0}&0.847 (0.156)&3.128 (7.216)&1.321 (3.720)&1 (2)&0.833 (0.333)\\[2pt]
        \textbf{StdUNet-DiceBCE-0.8}&0.844 (0.161)&3.128 (9.163)&1.394 (3.806)&1 (2)&0.800 (0.333)\\[2pt]
        \textbf{ResUNet-DiceBCE-1.0}&0.839 (0.156)&3.128 (8.682)&1.306 (3.667)&1 (2)&0.800 (0.333)\\[2pt]
        \textbf{ResUNet-UFL-1.0}&0.839 (0.169)&3.497 (6.961)&1.453 (4.372)&1 (2)&0.800 (0.333)\\[2pt]
        \textbf{StdUNet-UFL-0.8}&0.838 (0.168)&3.788 (9.037)&1.340 (4.503)&1 (2)&0.800 (0.333)\\[2pt]
        \textbf{StdUNet-UFL-1.0}&0.834 (0.170)&3.233 (7.648)&1.509 (4.441)&1 (2)&0.800 (0.333)\\[2pt]
        \hline
        \multicolumn{6}{p{0.9\linewidth}}{Values are expressed as median (interquartile range). The models are sorted according to median DSC. The suffix of each model name corresponds to the axial resolution of the input images (1.0×1.0×2.0 or 0.8×0.8×3.0 $mm^3$). Abbreviations $\rightarrow$ DSC: Dice similarity coefficient; HD95: 95th percentile Hausdorff distance; AVD: absolute volume difference; ALD: absolute lesion count difference; LF1: lesion detection F1 score.}
    \end{tabular}
    \label{tab:val_isla_submodels}
\end{table*}

\section{Model performance and inter-rater agreement on the double-annotated subset}
\label{appendix_inter_rater}
\setcounter{figure}{0}
Figure \ref{fig:inter_rater_metrics} shows the segmentation metrics for the 28 double-annotated cases from the external evaluation cohort.

\begin{figure*}
    \centering
    \begin{minipage}[c]{0.8\linewidth}
        \centering
        \includegraphics{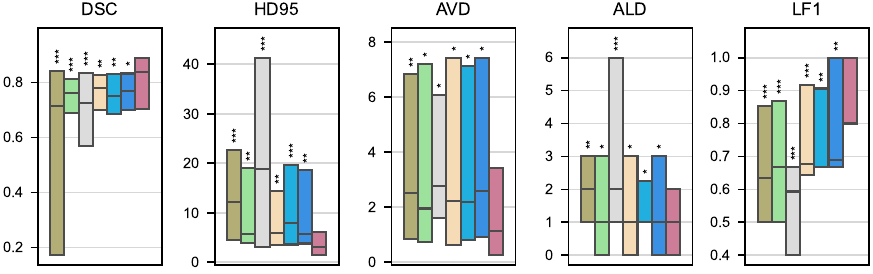}
    \end{minipage}
    \begin{minipage}[c]{0.17\linewidth}
        \centering
        \includegraphics{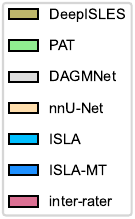}
    \end{minipage}
    \caption{\textbf{Segmentation metrics on the 28 double-annotated cases from the external evaluation cohort.} Asterisks indicate significant pairwise differences ($^{***}$: $p < 0.001$, $^{**}$: $p<0.01$, $^*$: $p < 0.05$) from the corresponding inter-rater metric. Boxplots display the median and interquartile range only; whiskers and outliers are omitted for clarity. Abbreviations $\rightarrow$ DSC: Dice similarity coefficient; HD95: 95th percentile Hausdorff distance; AVD: absolute volume difference; ALD: absolute lesion count difference; LF1: lesion detection F1 score.}
    \label{fig:inter_rater_metrics}
\end{figure*}

\section{Illustrative cases of the different categories of model errors}
\label{appendix_illustrative_cases}
\setcounter{figure}{0}
We provide illustrative cases of the different categories of model errors described in Section \ref{res:spatialErrors}: large FN produced by \textit{DeepISLES} (Figure \ref{fig:deepISLES_fn_examples}), FP produced by \textit{DAGMNet} in the temporal poles and anterior frontal regions (Figure \ref{fig:dagmnet_fp_temporalPoles_frontAnt_examples}), FP produced by \textit{DAGMNet} anterior to the temporal poles (Figure \ref{fig:dagmnet_fp_anteriorTemporalPoles_examples}), FP produced by \textit{PAT}, \textit{DAGMNet} and/or \textit{nnU-Net} in the posterior superior part of the brain (Figure \ref{fig:fp_patDagmnetNnunet_postSup_examples}), FN produced by \textit{PAT} and/or \textit{DAGMNet} in the deep sylvian region (Figure \ref{fig:fn_pat_dagmnet_deep_sylvian_examples}), FN produced by \textit{nnU-Net} and/or \textit{ISLA} in the deep sylvian region and detected by \textit{ISLA-MT} (Figure \ref{fig:fn_nnunet_isla_deep_sylvian_examples}), FN produced \textit{PAT}, \textit{DAGMNet} and/or \textit{nnU-Net} in the left superior region of the brain (Figure \ref{fig:fn_pat_dagmnet_nnunet_leftSup_examples}), FP produced by \textit{ISLA-MT} in the deep sylvian region (Figure \ref{fig:fp_islaMT_deep_sylvian_examples}).

\begin{figure*}
    \centering
    \begin{minipage}{\linewidth}
        \centering
        \includegraphics{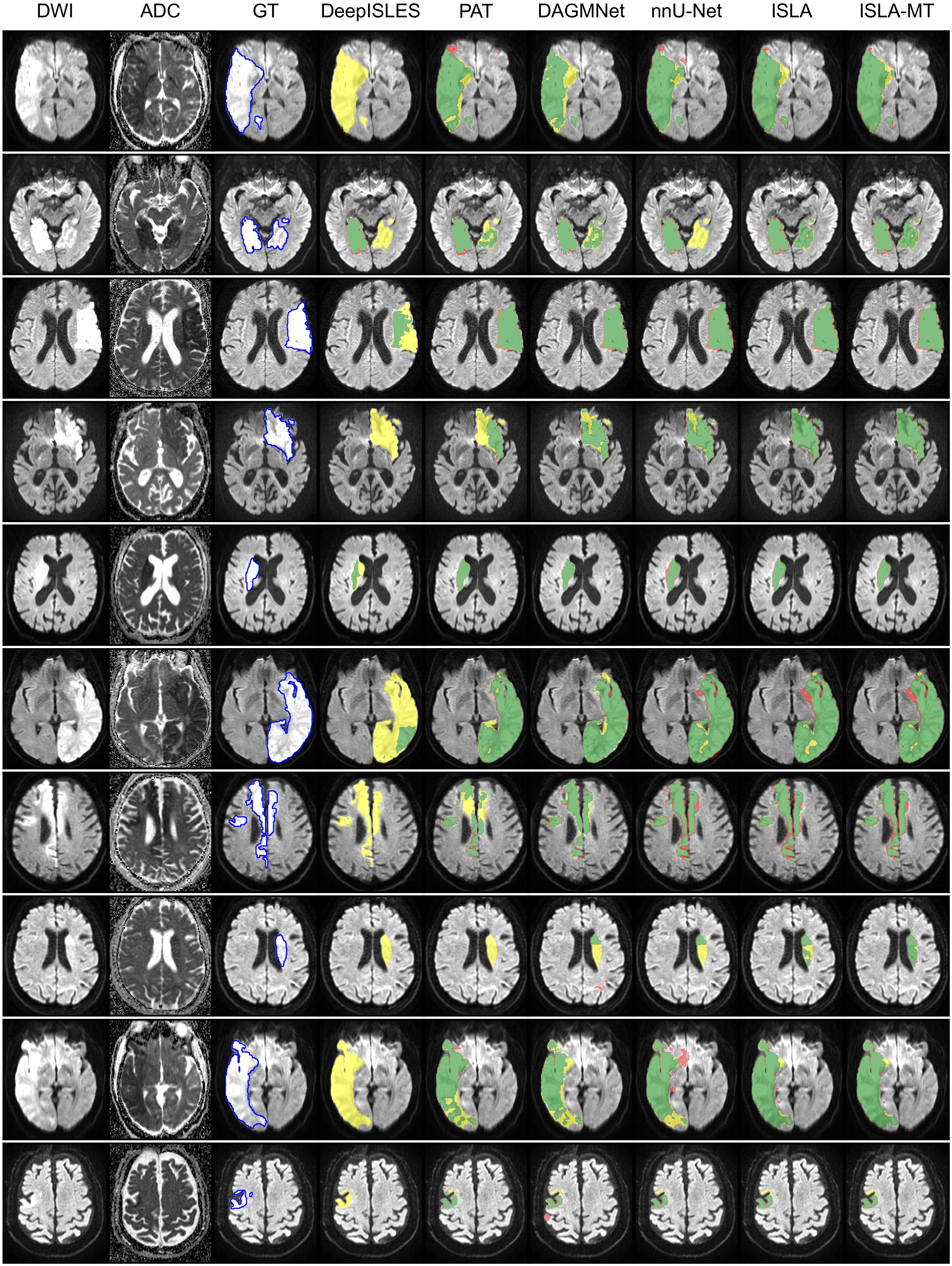}
    \end{minipage}
    \begin{minipage}{0.70\linewidth}
        \raggedleft
        \includegraphics{voxelClasses_legend.jpg}
    \end{minipage}
    \caption{\textbf{Illustrative examples of \textit{DeepISLES} false negatives across the brain.} Axial slices of the trace diffusion-weighted image (DWI), apparent diffusion coefficient map (ADC), and ground-truth annotations (GT) are shown alongside true-positive, false-negative, and false-positive voxels. All volumes were resampled to 0.8 × 0.8 × 3.0 mm$^3$ for visualization.}
    \label{fig:deepISLES_fn_examples}
\end{figure*}

\begin{figure*}
    \centering
    \begin{minipage}{\linewidth}
        \centering
        \includegraphics{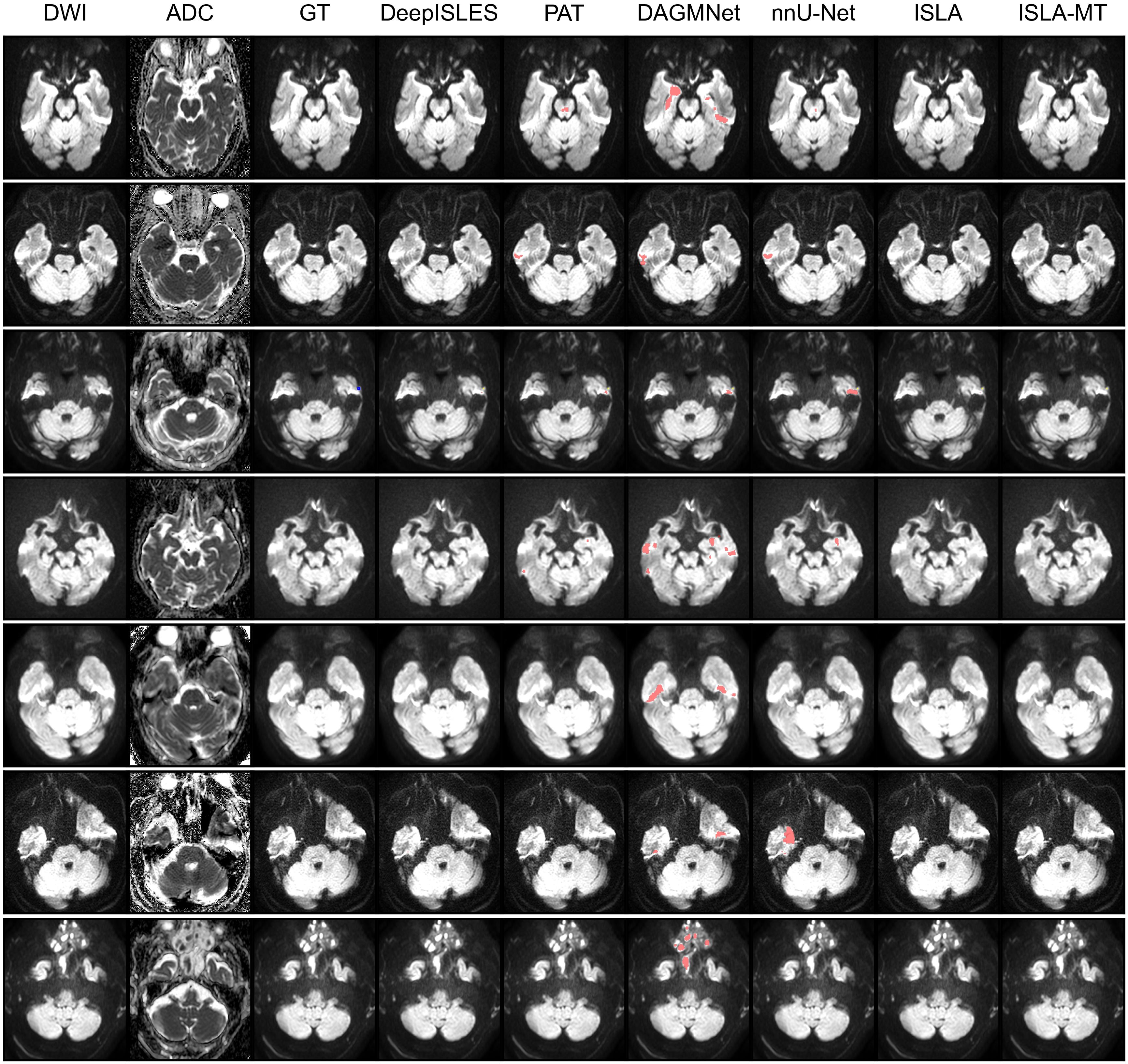}
    \end{minipage}
    \begin{minipage}{0.70\linewidth}
        \raggedleft
        \includegraphics{voxelClasses_legend.jpg}
    \end{minipage}
    \caption{\textbf{Illustrative examples of \textit{DAGMNet} false positives in the temporal poles and anterior frontal regions.} Axial slices of the trace diffusion-weighted image (DWI), apparent diffusion coefficient map (ADC), and ground-truth annotations (GT) are shown alongside true-positive, false-negative, and false-positive voxels. All volumes were resampled to 0.8 × 0.8 × 3.0 mm$^3$ for visualization.}
    \label{fig:dagmnet_fp_temporalPoles_frontAnt_examples}
\end{figure*}

\begin{figure*}
    \centering
    \begin{minipage}{\linewidth}
        \centering
        \includegraphics{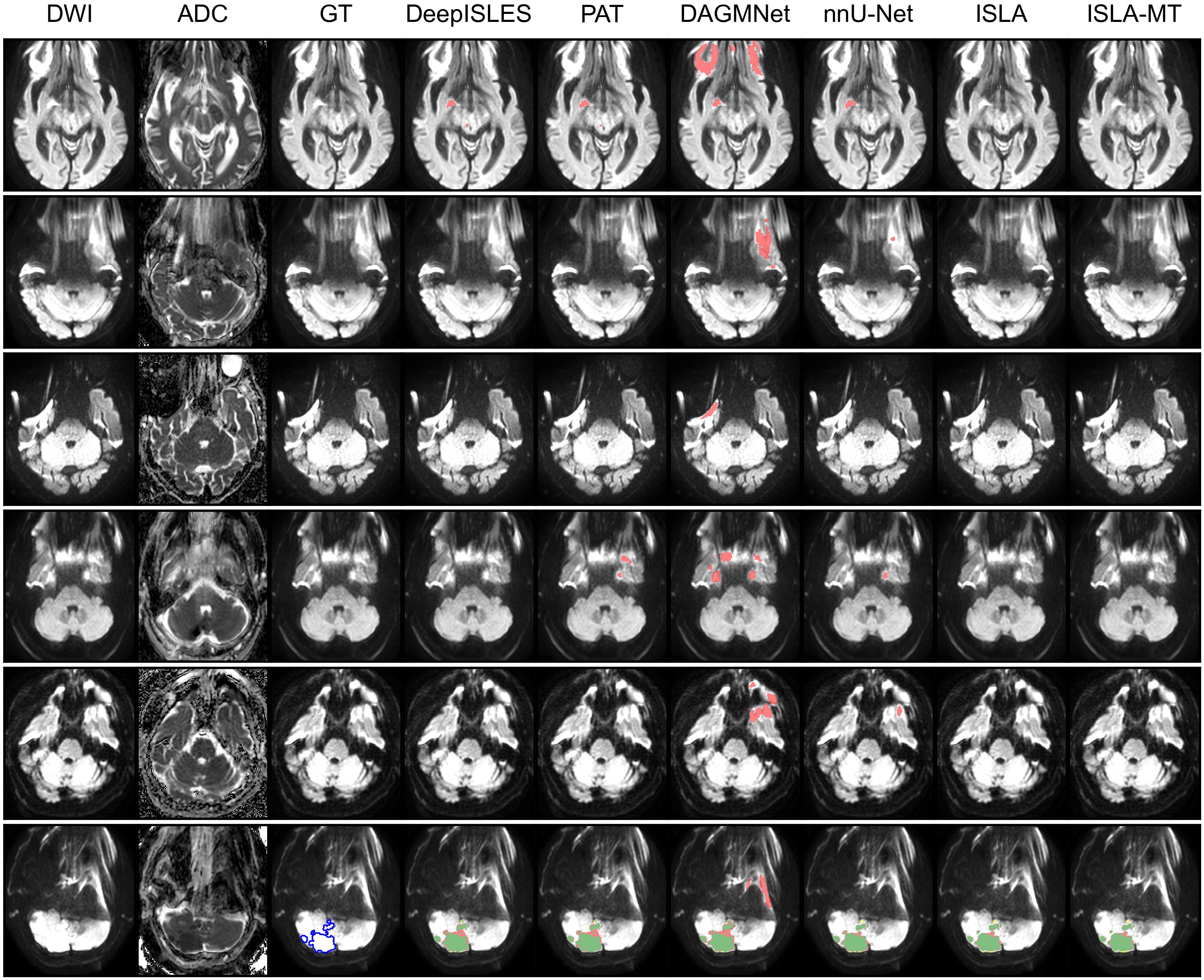}
    \end{minipage}
    \begin{minipage}{0.70\linewidth}
        \raggedleft
        \includegraphics{voxelClasses_legend.jpg}
    \end{minipage}
    \caption{\textbf{Illustrative examples of \textit{DAGMNet} false positives anterior to the temporal poles.} Axial slices of the trace diffusion-weighted image (DWI), apparent diffusion coefficient map (ADC), and ground-truth annotations (GT) are shown alongside true-positive, false-negative, and false-positive voxels. All volumes were resampled to 0.8 × 0.8 × 3.0 mm$^3$ for visualization.}
    \label{fig:dagmnet_fp_anteriorTemporalPoles_examples}
\end{figure*}

\begin{figure*}
    \centering
    \begin{minipage}{\linewidth}
        \centering
        \includegraphics{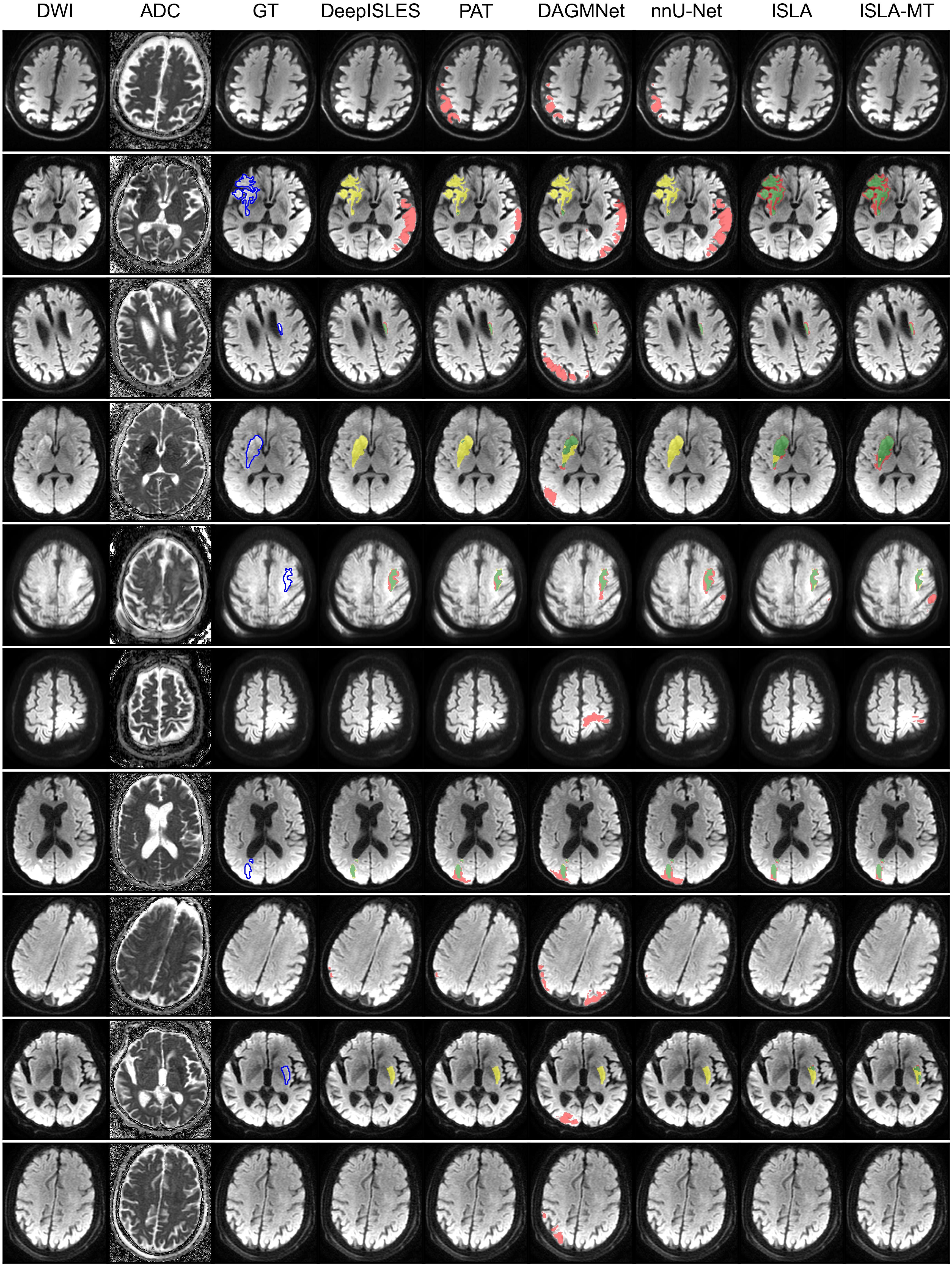}
    \end{minipage}
    \begin{minipage}{0.70\linewidth}
        \raggedleft
        \includegraphics{voxelClasses_legend.jpg}
    \end{minipage}
    \caption{\textbf{Illustrative examples of false positives produced by \textit{PAT}, \textit{DAGMNet} and/or \textit{nnU-Net} in the posterior superior part of the brain.} Axial slices of the trace diffusion-weighted image (DWI), apparent diffusion coefficient map (ADC), and ground-truth annotations (GT) are shown alongside true-positive, false-negative, and false-positive voxels. All volumes were resampled to 0.8 × 0.8 × 3.0 mm$^3$ for visualization.}
    \label{fig:fp_patDagmnetNnunet_postSup_examples}
\end{figure*}

\begin{figure*}
    \centering
    \begin{minipage}{\linewidth}
        \centering
        \includegraphics{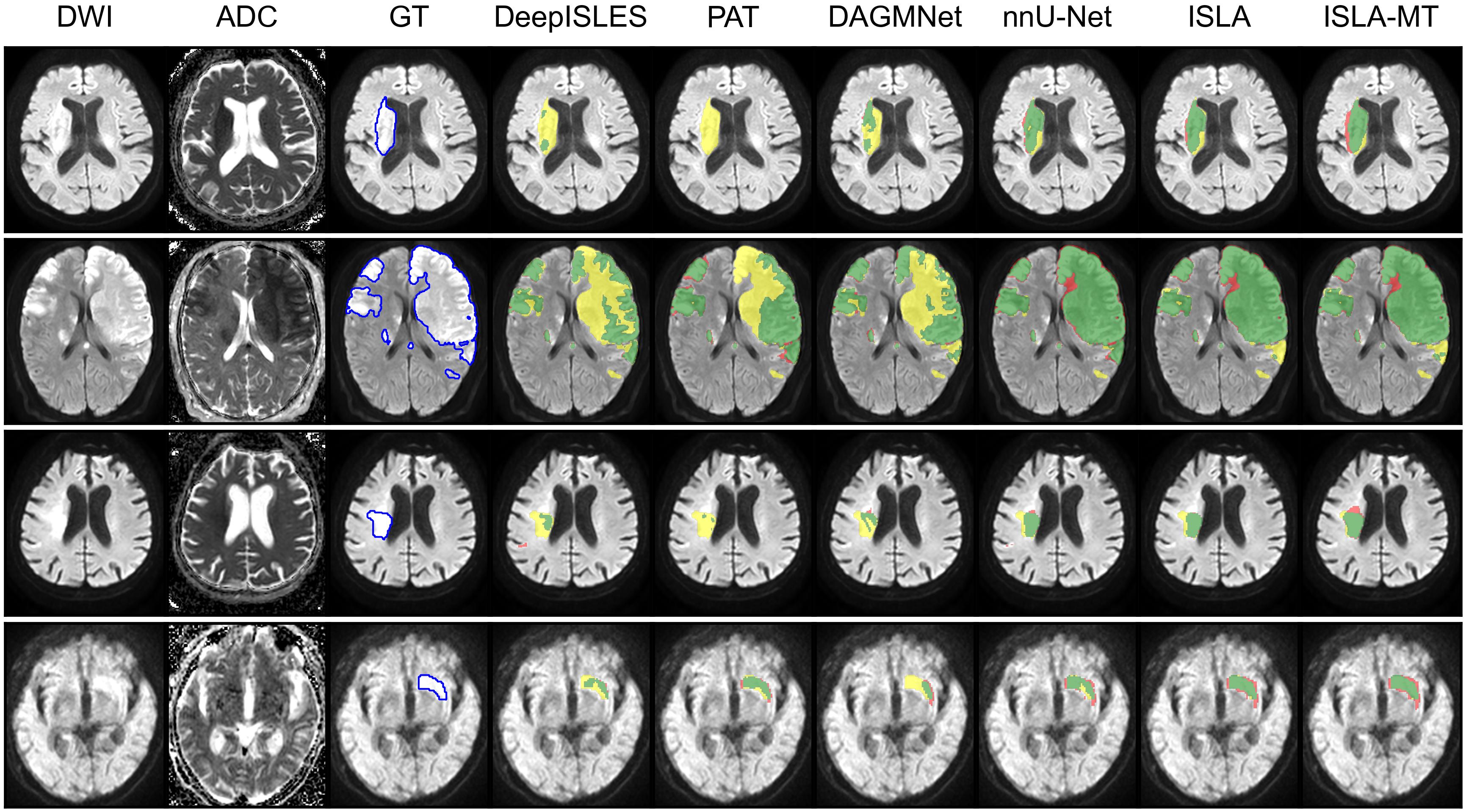}
    \end{minipage}
    \begin{minipage}{0.70\linewidth}
        \raggedleft
        \includegraphics{voxelClasses_legend.jpg}
    \end{minipage}
    \caption{\textbf{Illustrative examples of false negatives produced by \textit{PAT} and/or \textit{DAGMNet} in the deep sylvian region.} Axial slices of the trace diffusion-weighted image (DWI), apparent diffusion coefficient map (ADC), and ground-truth annotations (GT) are shown alongside true-positive, false-negative, and false-positive voxels. All volumes were resampled to 0.8 × 0.8 × 3.0 mm$^3$ for visualization.}
    \label{fig:fn_pat_dagmnet_deep_sylvian_examples}
\end{figure*}

\begin{figure*}
    \centering
    \begin{minipage}{\linewidth}
        \centering
        \includegraphics{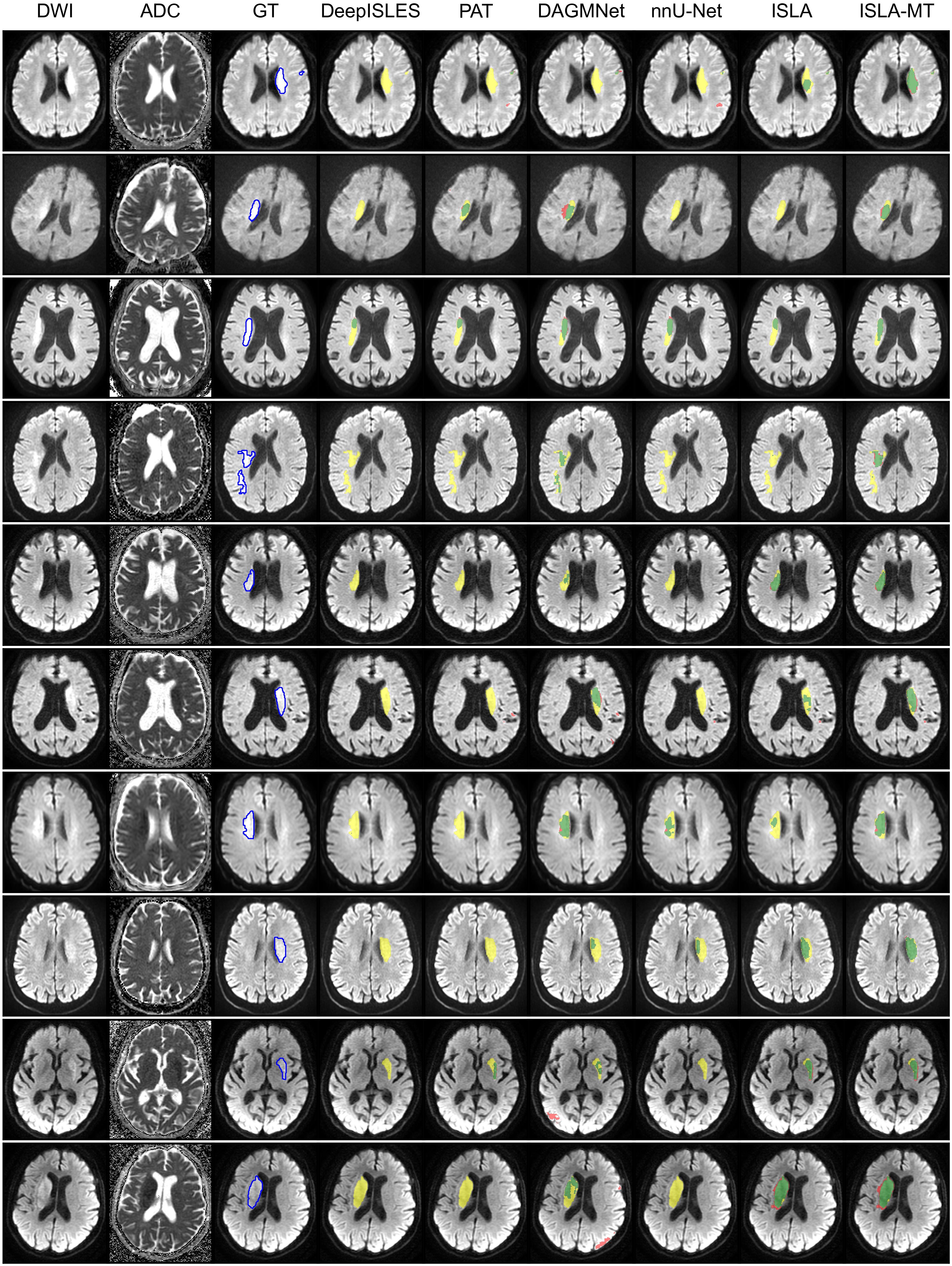}
    \end{minipage}
    \begin{minipage}{0.70\linewidth}
        \raggedleft
        \includegraphics{voxelClasses_legend.jpg}
    \end{minipage}
    \caption{\textbf{Illustrative examples of false negatives produced by \textit{nnU-Net} and/or \textit{ISLA} in the deep sylvian region and detected by \textit{ISLA-MT}.} Axial slices of the trace diffusion-weighted image (DWI), apparent diffusion coefficient map (ADC), and ground-truth annotations (GT) are shown alongside true-positive, false-negative, and false-positive voxels. All volumes were resampled to 0.8 × 0.8 × 3.0 mm$^3$ for visualization.}
    \label{fig:fn_nnunet_isla_deep_sylvian_examples}
\end{figure*}

\begin{figure*}
    \centering
    \begin{minipage}{\linewidth}
        \centering
        \includegraphics{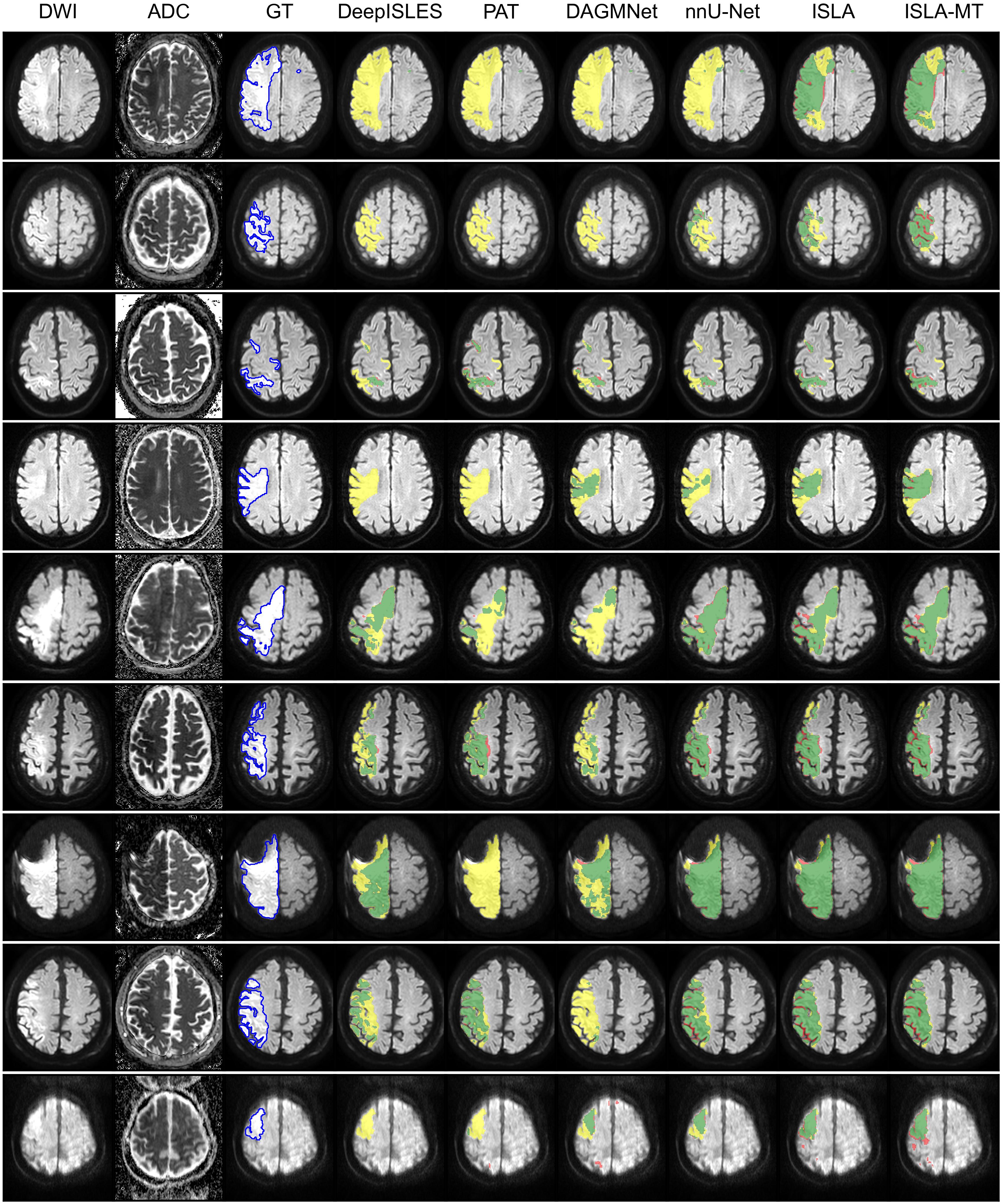}
    \end{minipage}
    \begin{minipage}{0.70\linewidth}
        \raggedleft
        \includegraphics{voxelClasses_legend.jpg}
    \end{minipage}
    \caption{\textbf{Illustrative examples of false negatives produced by \textit{PAT}, \textit{DAGMNet} and/or \textit{nnU-Net} in the left superior region of the brain.} Axial slices of the trace diffusion-weighted image (DWI), apparent diffusion coefficient map (ADC), and ground-truth annotations (GT) are shown alongside true-positive, false-negative, and false-positive voxels. All volumes were resampled to 0.8 × 0.8 × 3.0 mm$^3$ for visualization.}
    \label{fig:fn_pat_dagmnet_nnunet_leftSup_examples}
\end{figure*}

\begin{figure*}
    \centering
    \begin{minipage}{\linewidth}
        \centering
        \includegraphics{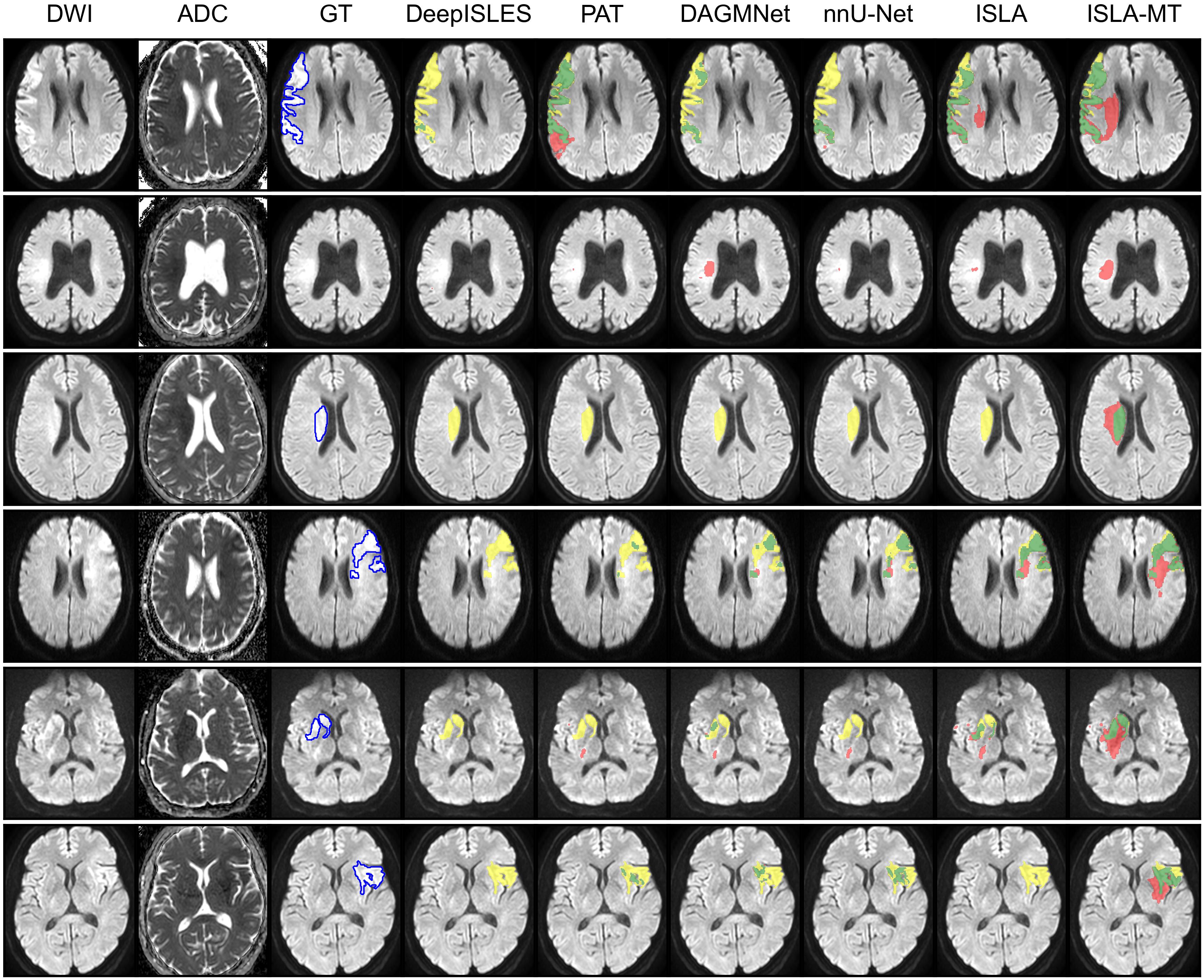}
    \end{minipage}
    \begin{minipage}{0.70\linewidth}
        \raggedleft
        \includegraphics{voxelClasses_legend.jpg}
    \end{minipage}
    \caption{\textbf{Illustrative examples of false positives produced by \textit{ISLA-MT} in the deep sylvian region.} Axial slices of the trace diffusion-weighted image (DWI), apparent diffusion coefficient map (ADC), and ground-truth annotations (GT) are shown alongside true-positive, false-negative, and false-positive voxels. All volumes were resampled to 0.8 × 0.8 × 3.0 mm$^3$ for visualization.}
    \label{fig:fp_islaMT_deep_sylvian_examples}
\end{figure*}

\section{Additional segmentation metrics}
\label{appendix_supplementaryMetrics}
\setcounter{table}{0}
Table \ref{tab:supplementaryMetrics} reports the median and interquartile range for DSC, precision, and recall across the external evaluation cohort.

\begin{table*}
    \centering
    \caption{\textbf{Dice similarity coefficient (DSC), precision and recall in the external evaluation cohort.}}
    \begin{tabular}{>{\raggedright}p{0.13\linewidth}>{\raggedright}p{0.11\linewidth}>{\raggedright}p{0.11\linewidth}>{\raggedright}p{0.11\linewidth}>{\raggedright}p{0.11\linewidth}>{\raggedright}p{0.11\linewidth}>{\raggedright\arraybackslash}p{0.11\linewidth}}
        \hline
        &\textbf{DeepISLES}&\textbf{PAT}&\textbf{DAGMNet}&\textbf{nnU-Net}&\textbf{ISLA}&\textbf{ISLA-MT}\\[2pt]
        \textbf{DSC}&0.644 (0.743)&0.762 (0.233)&0.735 (0.233)&0.765 (0.185)&0.772 (0.186)&0.781 (0.176)\\
        \textbf{precision}&0.865 (0.264)&0.768 (0.212)&0.749 (0.260)&0.740 (0.235)&0.764 (0.201)&0.743 (0.213)\\
        \textbf{recall}&0.606 (0.843)&0.820 (0.343)&0.804 (0.291)&0.880 (0.243)&0.860 (0.279)&0.891 (0.181)\\[2pt]
        \hline
        \multicolumn{7}{l}{Values are expressed as median (interquartile range).}
    \end{tabular}
    \label{tab:supplementaryMetrics}
\end{table*}

\section{Segmentation metrics of \textit{DeepISLES} and its constituent models}
\setcounter{table}{0}
\label{appendix_metric_deepISLES_subModels}
Table \ref{tab:metrics_deepISLES_subModels} reports the segmentation metrics of \textit{DeepISLES} and its constituent models, which are combined through majority voting \citep{delaRosa2025}. Two patients from the external evaluation cohort were excluded from this analysis because FLAIR images were unavailable, whereas they are required by the SWAN model.

\begin{table*}
    \centering
    \caption{\textbf{Segmentation metrics of \textit{DeepISLES} and its constituent models in the 588 participants with available T2-weighted fluid-attenuated inversion recovery images in the external evaluation cohort.}}
    \begin{tabular}{>{\raggedright}p{0.17\linewidth}>{\raggedright}p{0.14\linewidth}>{\raggedright}p{0.14\linewidth}>{\raggedright}p{0.14\linewidth}>{\raggedright}p{0.11\linewidth}>{\raggedright\arraybackslash}p{0.14\linewidth}}
    \hline
    &\textbf{DSC}&\textbf{HD95}&\textbf{AVD}&\textbf{ALD}&\textbf{LF1}\\[2pt]
    \textbf{DeepISLES}&0.645 (0.745)&12.153 (39.426)&5.210 (14.167)&1 (2)&0.667 (0.478)\\
    \textbf{SEALS}&0.552 (0.787)&19.321 (54.089)&6.385 (15.488)&1 (2)&0.667 (0.548)\\
    \textbf{NVAUTO}&0.679 (0.427)&11.146 (32.509)&6.486 (17.027)&3 (4)&0.545 (0.427)\\
    \textbf{SWAN}&0.595 (0.754)&14.290 (47.519)&5.216 (14.868)&2 (2)&0.667 (0.524)\\[2pt]
    \hline
    \multicolumn{6}{l}{Values are expressed as median (interquartile range).}
    \end{tabular}
    \label{tab:metrics_deepISLES_subModels}
\end{table*}

\section{Precision and recall in the ablation studies}
\label{appendix_ablation_precision_recall}
\setcounter{table}{0}
We reproduced the ablation experiments described in Section \ref{meth:ablations} using precision and recall as evaluation metrics. The results are reported in Table \ref{tab:ablation_precision_recall}.

\begin{table}
    \centering
    \caption{\textbf{Comparison of precision and recall between the ablated and the full \textit{ISLA} variants in the external evaluation cohort.}}
    \begin{tabular}{>{\raggedright}p{0.40\linewidth}>{\raggedright}p{0.22\linewidth}>{\raggedright\arraybackslash}p{0.22\linewidth}}
        \hline
        \textbf{ablated variant}&\textbf{precision}&\textbf{recall}\\[5pt]
        \hdashline
        &\multicolumn{2}{c}{\textbf{ISLA}}\\[2pt]
        
        \textbf{1.0×1.0×2.0mm$\bm{^3}$}&$+0.149^{**}$&$-0.469^{***}$\\[2pt]
        \textbf{0.8×0.8×3.0mm$\bm{^3}$}&$-0.225^{***}$&$+0.172^{***}$\\[4pt]
        
        \textbf{StdUNet}&$-0.102^{*}$&$-0.035^{}$\\[2pt]
        \textbf{ResUNet}&$-0.037^{}$&$-0.090^{}$\\[4pt]
        
        \textbf{DiceBCE}&$-0.867^{***}$&$+0.748^{***}$\\[2pt]
        \textbf{UFL}&$+0.837^{***}$&$-0.710^{***}$\\[5pt]
        \hdashline

        &\multicolumn{2}{c}{\textbf{ISLA-MT}}\\[2pt]

        \textbf{1.0×1.0×2.0mm$\bm{^3}$}&$+0.490^{***}$&$-0.788^{***}$\\[2pt]
        \textbf{0.8×0.8×3.0mm$\bm{^3}$}&$+0.383^{***}$&$-0.595^{***}$\\[4pt]
        
        \textbf{StdUNet}&$+0.582^{***}$&$-0.698^{***}$\\[2pt]
        \textbf{ResUNet}&$-0.915^{***}$&$+0.839^{***}$\\[4pt]
        
        \textbf{DiceBCE}&$-0.820^{***}$&$+0.699^{***}$\\[2pt]
        \textbf{UFL}&$+0.807^{***}$&$-0.813^{***}$\\[2pt]
        \hline
        \multicolumn{3}{p{0.95\linewidth}}{The name of each ablated variant indicates the component retained in the ensemble inference (e.g., only residual blocks in the ResUNet ablation). Rank-biserial correlation coefficients ($r_{rb}$) are reported, along with the corresponding significance levels ($^{***}$: $p < 0.001$, $^{**}$: $p < 0.01$, $^*$: $p < 0.05$). Negative and positive $r_{rb}$ values indicate a decrease and increase, respectively, in the corresponding metric due to ablation. Abbreviations $\rightarrow$ DSC: Dice similarity coefficient; HD95: 95th percentile Hausdorff distance; AVD: absolute volume difference; ALD: absolute lesion count difference; LF1: lesion detection F1 score.}
    \end{tabular}
    \label{tab:ablation_precision_recall}
\end{table}

\bibliographystyle{elsarticle-harv}\biboptions{authoryear}
\bibliography{refs}

\end{document}